\documentclass[final,5p,times]{elsarticle}

\usepackage{lineno,hyperref}
\modulolinenumbers[5]

\journal{Robotics and Autonomous Systems}

\usepackage{amsfonts,amssymb}
\usepackage{amsmath}
\usepackage{bm}
\usepackage{graphicx}
\usepackage{color}
\usepackage[ruled,vlined,commentsnumbered]{algorithm2e}
\usepackage{amsthm}
\usepackage{comment}

\usepackage{multicol}
\usepackage{float}
\usepackage{tabularx,booktabs}
\usepackage[table]{xcolor}
\usepackage{tikz}

\usepackage{mwe}
\usepackage{lettrine}
\usepackage{xcolor}

\newtheorem{definition}{Definition}

\newtheorem{remark}{Remark}

\newtheorem{problem}{Problem}
\newtheorem{assumption}{Assumption}

\begin{document}

\begin{frontmatter}

\title{Specification mining and automated task planning for autonomous robots based on a graph-based spatial temporal logic
\tnoteref{mytitlenote}}
\tnotetext[mytitlenote]{Work carried out whilst at University of Notre Dame}



\author{Zhiyu Liu\corref{cor1}\fnref{label1}}
\author{Meng Jiang\fnref{label2}}
\author{Hai Lin\fnref{label1}}

\cortext[cor1]{Corresponding author}
\fntext[label1]{Department of Electrical Engineering, University of Notre Dame}
\fntext[label2]{Department of Computer Science and Engineering, University of Notre Dame}

\begin{abstract}
We aim to enable an autonomous robot to learn new skills from demo videos and use these newly learned skills to accomplish non-trivial high-level tasks. The goal of developing such autonomous robot involves knowledge representation, specification mining, and automated task planning. For knowledge representation, we use a graph-based spatial temporal logic (GSTL) to capture spatial and temporal information of related skills demonstrated by demo videos. We design a specification mining algorithm to generate a set of parametric GSTL formulas from demo videos by inductively constructing spatial terms and temporal formulas.
The resulting parametric GSTL formulas from specification mining serve as a domain theory, which is used in automated task planning for autonomous robots. We propose an automatic task planning based on GSTL where a proposer is used to generate ordered actions, and a verifier is used to generate executable task plans. A table setting example is used throughout the paper to illustrate the main ideas. 
\end{abstract}

\begin{keyword}
Spatial temporal logic \sep Knowledge representation \sep Specification mining \sep Automated task planning
\end{keyword}

\end{frontmatter}

%\linenumbers

\section{Introduction}

Our work is motivated by the quest for high-level autonomy in service robots that can adapt to our everyday life. The key challenge comes from the fact that we cannot pre-program the robot since the working environment of a service robot is unpredictable \cite{paulius2019survey}, and the tasks for the robot to accomplish could be new in the sense that the robot has never been trained before. Instead of waiting for hours' trial-and-error, we expect the robot can learn new skills instantly and achieve high-level tasks even when the tasks are only partially or vaguely specified, e.g., ``John is coming for dinner, set-up the dinner table.'' Setting up a table for dinner may be new for the robot if it has never done the task before. 

%In the last two decades, there are great developments in areas related to robotics such as sensors technology \cite{yousef2011tactile}, cloud computing \cite{sheta2019robotics}, and artificial intelligence \cite{hernandez2017evaluation}. Robots are more capable than ever to achieve fully autonomy leading to great research interests in employing autonomous robots in applications such as service robots \cite{liu2018coordinated}, manufacturing \cite{zheng2019vector}, and search and rescue \cite{rashid2019collabdrone}. In these applications, pre-programmed robots are able to execute certain tasks under a well defined and constrained environment. However, robots still cannot achieve fully autonomy which requires robots to operate in real-world for weeks or months without human interference. One of the biggest challenges for autonomous robot is its lack of ability to learn and reason for its own \cite{kazakov2001machine}. Specifically, since the real world can defy even our wildest imaginations, there will always be unexpected situations that designers fail to consider for robots. This requires robots to be able to constantly learn new knowledge by observing the environment. Furthermore, robots will often have to deal with an incomplete or even vague task assignment, which requires robots to complete the task assignment with detailed executable plans. The challenges motivate us to develop autonomous robots with the ability to make independent decisions in the real world through actively learning from the environment and automatically generating an executable task plan. 

The solution we propose is to learn related new skills through observing demo videos and apply the skills in solving a vague task assignment. In the solution, we assume that the robot can go to internet and fetch demo videos for the task at hand (like we learn new skills by searching Google or watching YouTube videos). We also assume that the robot can reliably detect the objects in the demo videos and match the objects in its surrounding environment. Our key idea of the solution is to formally specify the learned skills by employing a graph-based spatial temporal logic (GSTL), which was proposed recently for knowledge representation in autonomous robots \cite{liu2020graph}. GSTL enables us to formally represent both spatial and temporal knowledge that is essential for autonomous robots. It is also shown in \cite{liu2020graph} that the satisfiability problem in GSTL is decidable and can be solved efficiently by SAT.
In this paper, we further ask (a) how to automatically mine GSTL specifications from demo videos and generate a domain theory, and (b) how to achieve an automated task planning based on the newly learned domain theory. 

Specifically, for the first question, we propose a new specification mining algorithm that can learn a set of parametric GSTL formulas describing spatial and temporal relations from the video. By parametric GSTL formulas, we mean the temporal and spatial variables in GSTL formulas are yet to be decided. 
Specification mining for spatial logic or temporal logic has been studied separately in the literature \cite{kong2017temporal,nenzi2018robust,bombara2016decision,jin2015mining,bartocci2016formal}. However, we cannot simply combine the existing specification mining techniques for GSTL as GSTL has a broader expressiveness (e.g. parthood, connectivity, and metric extension) compared to existing spatial temporal logic \cite{liu2020graph}.
The present techniques face difficulties when it models actions involving coupled spatial and temporal information with metric extension, e.g., a hand holds a plate with a cup on top of it for one minute. To handle this difficulty, our basic idea is to generate simple spatial GSTL terms and construct more complicated spatial terms and temporal formulas based on the simple spatial terms inductively. Specifically, we construct spatial terms by mining both parthood and connectivity of the spatial elements and temporal formulas by considering preconditions and consequences of a given spatial term. The obtained parametric GSTL formulas can represent skills demonstrated by the video and form as a domain theory to facilitate automated task planning.

For the second question, we propose an interacting proposer and verifier to achieve an automated task planning based on the newly learned domain theory in parametric GSTL.  
Many approaches have been proposed in AI and robotics on automated task planning, which can be roughly divided into several groups, including graph searching \cite{weld1999recent}, model checking \cite{li2012planning}, dynamic programming \cite{zhou2018mobile}, and MDP \cite{zheng2019vector}. Despite the success of existing approaches, there is a big gap between planning and executing where the task plan cannot guarantee the feasibility of the plans. 
The interaction between proposer and verifier aims to fill this gap. The proposer generates ordered actions, and the verifier makes sure the plan is feasible and can be achieved by robots. In the proposer, we use the available actions in the domain theory as basic building blocks and generate ordered actions for the verifier. The verifier checks temporal and spatial constraints posed by the domain theory and sensors by solving an SMT satisfiability problem for the temporal constraints and an SAT satisfiability problem for the spatial constraints.

The contributions of this paper are mainly twofold. 
First, we propose a new specification mining algorithm for GSTL, where a set of parametric GSTL is learned from demo videos. The proposed specification mining algorithm can mine both spatial and temporal information from the video with limited data. The parametric GSTL formulas form the domain theory, which is used in the planning. Our work differs from the existing work where the domain theory is assumed to be given. 
Second, we design an automatic task planning framework containing an interactive proposer and a verifier for autonomous robots. The proposer can generate ordered actions based on the domain theory and the task assignment. The verifier can verify the feasibility of the proposed task plan and generate time instances for an executable action plan. The overall framework is able to independently solve a vague task assignment with detailed and executable action plans with limited human inputs. 

The rest of the paper is organized as follows. In Section \ref{related work}, we introduce related work on pursuing autonomous robots. In Section \ref{problem formulation and assumption}, we give a motivating scenario and formally state the problem. In Section \ref{Section: GSTL definitio}, we briefly introduce the graph-based spatial temporal logic, GSTL. In Section \ref{section:specification mining}, we introduce the specification mining algorithm based on demo videos. An automatic task planning framework is given in Section \ref{section:task planning}. We evaluate the proposed algorithms in Section \ref{section:evaluation} with a table setting example. Section \ref{Section: conclusion} concludes the paper.

\section{Related work}\label{related work}

\begin{comment}

\subsection{Autonomous robots}
The journey of achieving autonomy for autonomous robots starts at sensing and modeling the changing environment. This goal has been achieved through the development of sensors and extensive research on environment modeling such as SLAM \cite{khairuddin2015review}. Current research on autonomous robots focuses on specialization of robots by improving both general robustness and specific task performance \cite{kunze2018introduction}. Robots with various degrees of autonomy have been applied in space \cite{oettershagen2017design}, field \cite{broggi2012vislab,bechar2016agricultural}, and service \cite{hanheide2017and,tran2017robots,decker2017service}. Nearly all applications include the following key components, namely navigation $\&$ mapping, perception, knowledge representation $\&$ reasoning, task planning, and learning \cite{kunze2018artificial}. Each component is supported differently depending on applications. Navigation $\&$ mapping
and perception are well supported in the areas mentioned above while knowledge representation $\&$ reasoning, task planning, and learning are only partially supported. However, many of them are strictly limited in specific task assignment which leaves many challenges and research questions.

In long term, autonomous robots are required to adapt to the changing environment through learning and knowledge representation and reasoning. We introduce related work on knowledge representation and reasoning, learning, and automated task planning based on them in the following sections.

\end{comment}

A high-level autonomy for mobile robots is a very ambitious goal that needs support from many areas, such as navigation $\&$ mapping, perception, knowledge representation $\&$ reasoning, task planning, and learning. In this section, we will briefly introduce the most relevant work to us in knowledge representation, specification mining, and automated task planning.

\subsection{Knowledge representation and reasoning} 
One of the most promising fields in knowledge representation and reasoning is the logic-based approach, where knowledge is modeled by predefined elementary (logic and non-logic) symbols \cite{wachter2018integrating}, and automated planning is performed through primitive operations manipulating the symbols \cite{hertzberg2008ai}. Classic logic such as propositional logic \cite{post1921introduction}, first-order logic \cite{mccarthy1960programs}, and description logic \cite{baader2003description} are well developed and can be used to represent knowledge with a great expressiveness power in different domains. However, this is achieved at the expense of tractability. The satisfiability problem of classic logic is often undecidable, which further limits its application in autonomous robots. Furthermore, in general, classic logic fails to capture the temporal and spatial characteristics of the knowledge. For example, it is difficult to capture information such as a robot hand is required to hold a cup for at least five minutes. As spatial and temporal information are often particularly important for autonomous robots, spatial logic and temporal logic are studied both separately \cite{cohn2001qualitative,raman2015reactive} and combined \cite{kontchakov2007spatial,haghighi2016robotic,bartocci2017monitoring,liu2020graph}. By integrating spatial and temporal operators with classic logic operators, spatial temporal logic shows great potential in specifying a wide range of task assignments for autonomous robots with automated reasoning ability. 

However, two significant concerns limit the applications of spatial temporal logic in autonomous robots. First, the knowledge needed for task planning is often given by human experts in advance. The dependence on human experts is caused by the lack of specification mining algorithm between the real world and the symbolic-based knowledge representation. 
%The sensing data from the real world is often continuous and noisy, while the symbolic-based knowledge representation and reasoning operate on discretized symbols. This requires us to interpret the sensor data stream on a semantic level to transform it into a symbolic description in terms of categories that are meaningful in the knowledge representation \cite{hertzberg2008ai}. Most related works lack a specification mining block to fulfill this role and simply give the knowledge in advance. 
Second, there are few results in spatial temporal logic where the task plan is automatically generated and is executable and explainable to robots. Lots of existing spatial temporal logic are undecidable due to their combination of spatial operators and temporal operators \cite{kontchakov2007spatial}, and the resulting task plan may not be feasible for robots to complete. 
In summary, the lack of specification mining and executable task planning limits spatial temporal logic's applications on autonomous robots.

\subsection{Learning and specification mining}
Learning is essential for autonomous robots since deployment in real worlds with considerable uncertainty means any knowledge the robot has is unlikely to be sufficient. By focusing on specific scenarios, robots can increase their knowledge through learning, which has been applied to applications such as assembling robots \cite{suomalainen2017geometric} and service robots \cite{decker2017service}. As the applications in autonomous robots are often task-oriented, the goal of learning is often to find a set of control policies for given tasks. 
Such policy can be learned through approaches such as learning from demonstration \cite{argall2009survey} and reinforcement learning \cite{tsurumine2019deep,sutton2018reinforcement}. 

%In general, the goal of specification mining is to find a set of control policy for given states, where the policy can be modeled as low-level continuous trajectories or symbolic encoding for motion primitives. The approaches of generating such policy can be categorized into three group, namely mapping function, system model, and plans \cite{argall2009survey}. The mapping function approach calculates a function mapping states to actions. The basic idea is first mapping states to dynamic motion primitives (DMPs) using classifiers such as Bayesian network \cite{inoue1999acquisition} then the DMPs are ordered via Hidden Markov Models \cite{hovland1996skill}. Similar to reinforcement learning, the system model approach first builds a transition system of the world and derives control policy from it by maximising a reward function, which is normally given by human experts \cite{thomaz2006reinforcement}. The plans approach describes policy as a sequence of actions whose preconditions and effects are learned from examples \cite{kuniyoshi1994learning}.  

In a logic-based approach, learning is often addressed by specification mining, where a set of logic formulas are learned from data or examples. Most of the recent research has focused on the estimation of parameters associated with a given logic structure \cite{asarin2011parametric,bartocci2013robustness,jin2015mining,yang2012querying}. However, the selected formula may not reflect achievable behaviors or may exclude fundamental behaviors of the system. Furthermore, by giving the formula structure \emph{a prior}, the mining procedure cannot derive new knowledge from the data. Few approaches such as directed acyclic graph \cite{kong2017temporal} and decision tree \cite{bombara2016decision} are explored for temporal logic where the structures are not entirely fixed. 

Despite the success of specification mining, the majority of specification mining algorithms developed to date generate a purely reactive policy that maps directly from a state to action without considering temporal relations among actions \cite{argall2009survey}. They have difficulty addressing complicated temporal and spatial requirements, such as accomplishing particular task infinity often or holding a cup for at least 5 minutes. One possible solution is encoding temporal and spatial information in the policy derivation process. 

With the ability of learning and reasoning, robots can independently solve a task assignment through automated task planning. We review related work on automated task planning as follows.

\subsection{Automated task planning}
The automated task planning determines the sequence of actions to achieve the task assignment. Existing planning approaches such as GRAPHPLAN \cite{weld1999recent}, STRANDS \cite{hawes2017strands}, CoBot \cite{veloso2012cobots} and Tangy \cite{tran2017robots} are able to generate ordered actions for a given task assignment from users. Different systems vary on if they consider preconditions and effects of robots' actions on time and resources. In GRAPHPLAN, both preconditions and effects of actions are modeled during the task planning. In STRANDS and CoBot, task planning is generated based on models (e.g., MDP model for a working environment) learned from the previous execution. Even though existing work can generate ordered actions, there is no guarantee that the ordered actions are feasible and executable for robots when spatial and temporal constraints are considered. A verification process is needed for the ordered actions \cite{kunze2018artificial}.

As robots are more adaptable to a structured environment, the research trending is integrating learning and task planning processes in robots for a less structured environment. The performance of robots is evaluated over variation in task assignment and available resources \cite{kunze2018artificial} so that the plans are feasible and executable for robots.

\section{Problem formulation}\label{problem formulation and assumption}
\begin{figure}
 \centering
 \includegraphics[scale=0.54]{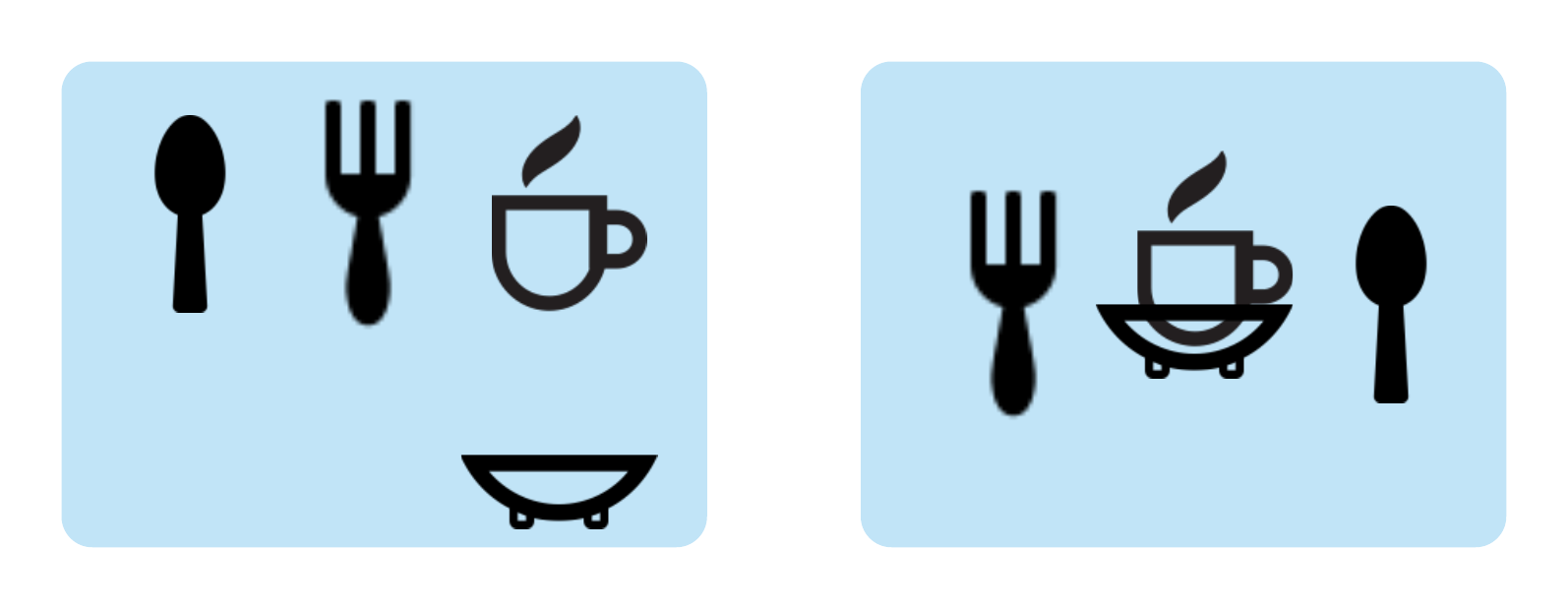}
 \caption{An example of specification mining and automatic task planning for the dining table setting. The left figure is the initial table setup. The right figure is the target table setup.}
 \label{exp:task planning example}
\end{figure}

We introduce the motivating scenarios and a formal problem statement in this section. In this paper, we aim to develop an autonomous robot with the ability to learn available actions from examples and generating executable actions to fulfill a given task assignment. The motivating scenario is given in Fig. \ref{exp:task planning example}, where the initial table setup is given in the left figure, and the goal is to set up the dining table as shown in the right figure. We formally state the problem as follows.
\begin{problem}
Given a set of demo videos $\mathcal{G}$ and an initial table setup $s_1,s_2,...$ as shown in the left in Fig. \ref{exp:task planning example}, we aim to accomplish a target table setup $s^*_1,s^*_2,...$ shown in the right in Fig. \ref{exp:task planning example} through an executable task plan $\psi$ as GSTL formulas. Here, $s_i$ and $s^*_i$ are GSTL spatial terms representing table setup. The problem is solved by solving the following two sub-problems.
\begin{enumerate}
    \item Generate a domain theory $\Sigma=\{a_1,a_2,...\}$ in GSTL via specification mining based on video $\mathcal{G}$.
    \item Generate the task plan $\psi$ based on the initial setup $s_i$, the target setup $s^*_i$, and the domain theory $\Sigma$.
\end{enumerate}
\label{problem statement}
\end{problem}

To solve Problem \ref{problem statement}, we adopt the following assumptions with justifications. First, we assume that we know the objects and concepts we are interested in and all the parthood relations for the objects. For example, we know ``hand is part of body part" and ``cup is a type of tool." Second, we assume reliable object detection with an accurate position tracking algorithm is available since there are many mature object detection algorithms \cite{zhao2019object} and stereo cameras like ZED can provide an accurate 3D position for objects \cite{chaudhary2018learning}.

\section{Graph-based spatial temporal logic}\label{Section: GSTL definitio}

In this section, we briefly introduce the graph-based spatial temporal logic (GSTL). First, we introduce the temporal and spatial representation for GSTL.

\subsection{Temporal and spatial representations}

There are multiple ways to represent time, e.g., continuous-time, discrete-time, and interval. As people are more interested in time intervals in autonomous robots, in this paper, we use a discrete-time interval. We employ Allen interval algebra (IA) \cite{allen1983maintaining} to model the temporal relations between two intervals. Allen interval algebra defines the following 13 temporal relationships between two intervals, namely before ($b$), meet ($m$), overlap ($o$), start ($s$), finish ($f$), during ($d$), equal ($e$), and their inverse ($^{-1}$) except equal. 

\begin{figure}[H]
    \centering
    \includegraphics[scale=0.3]{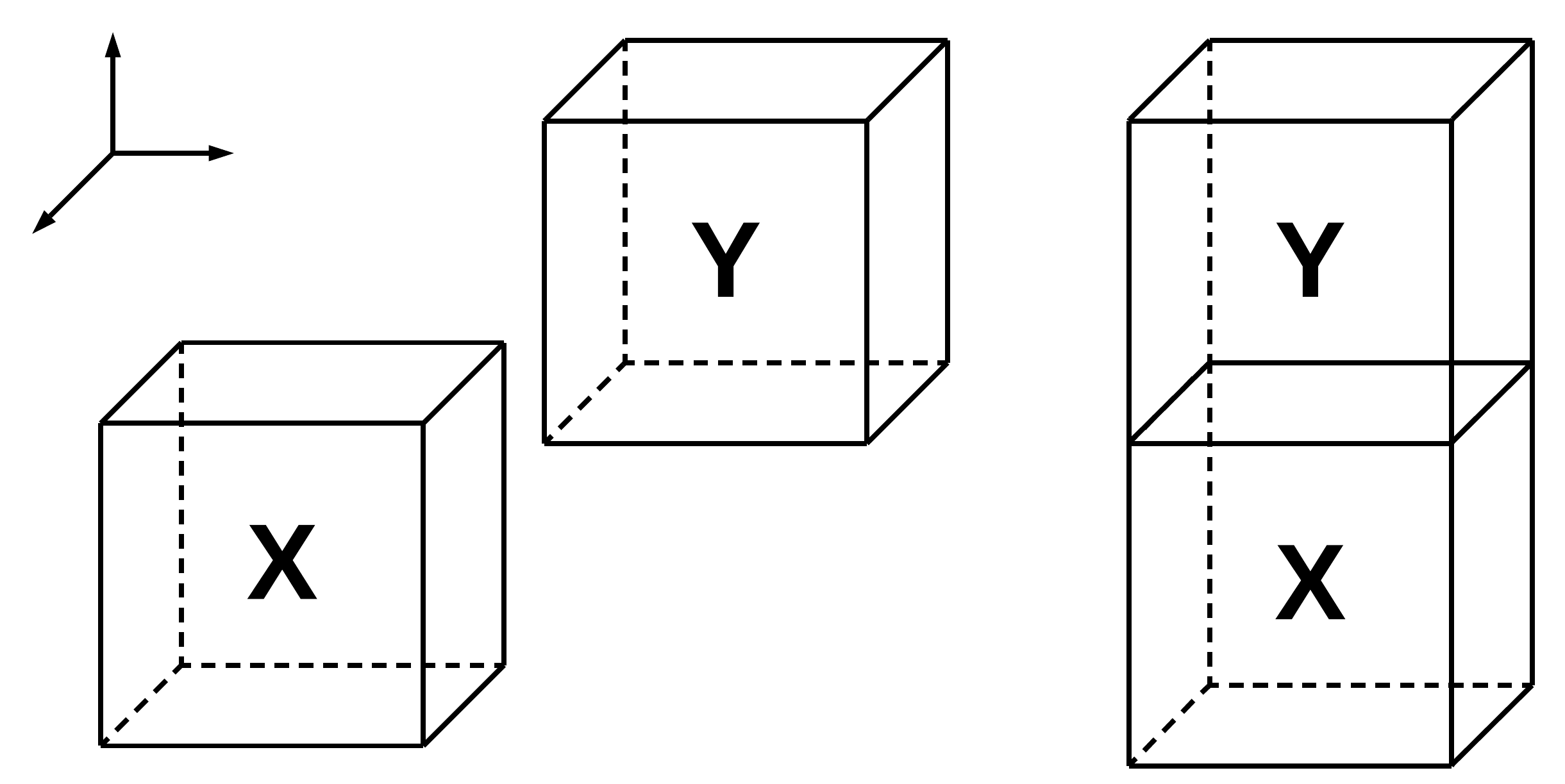}
    \caption{Representing directional relations between objects $X$ and $Y$ in 3D interval algebra}
    \label{Relations in CA}
\end{figure}

As for the spatial representations, we use regions as the basic spatial elements instead of points. Within the qualitative spatial representation community, there is a strong tendency to take regions of space as the primitive spatial entity \cite{cohn2001qualitative}. In practice, a reasonable constraint to impose would be that regions are all rational polygons. To consider the relations between regions, we consider parthood and connectivity in our spatial model, where parthood describes the relational quality of being a part, and connectivity describes if two spatial objects are connected. GSTL further includes directional information in connectivity. It is done by extending Allen interval algebra into 3D, which is more suitable for autonomous robots. The relations between two spatial regions are defined as $\mathcal{R}=\{(A,B,C):A,B,C\in \mathcal{R}_{IA}\}$, where $13\times 13\times 13$ basic relations are defined. An example is given in Fig. \ref{Relations in CA} to illustrate the spatial relations. For spatial objects, $X$ and $Y$ in the left where $X$ is at the front, left, and below of $Y$, we have $X\{(b,b,o)\}Y$. For spatial objects, $X$ and $Y$ in the right where $Y$ is completely on top of $X$, we have $X\{(e,e,m)\}Y$.

\begin{figure}
 \centering
 \includegraphics[scale=0.5]{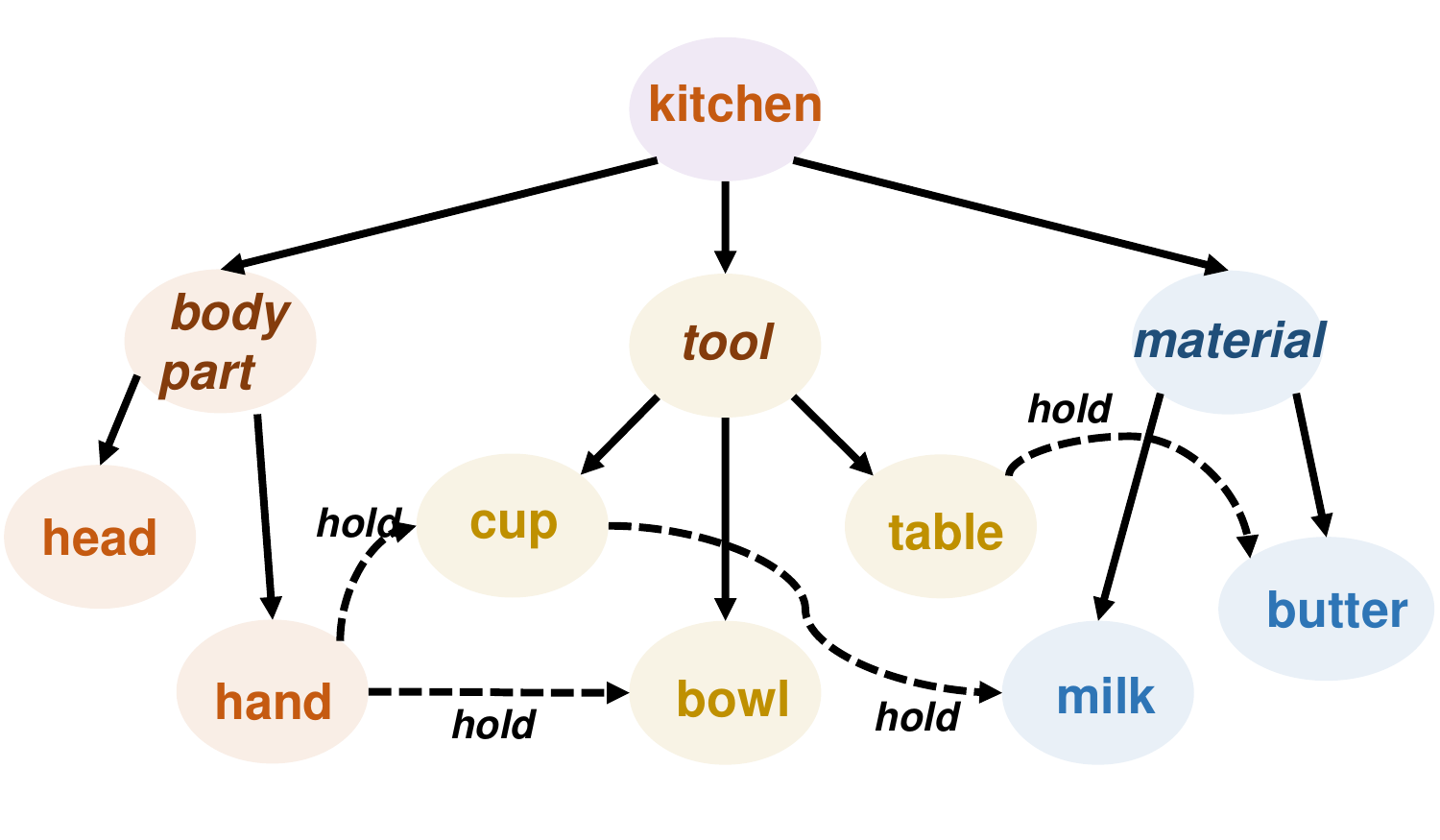}
 \caption{The hierarchical graph with three basic spatial operators: parent, child, and neighbor where the parent-child relations are drawn in solid line and the neighbor relations are drawn in dashed lines.}
 \label{exp:Graph with a hierarchy structure}
\end{figure}

We apply a graph with a hierarchy structure to represent the spatial model.
Denote $\Omega=\cup_{i=1}^{n}\Omega_i$ as the union of the sets of all possible spatial objects where $\Omega_i$ represents a certain set of spatial objects or concepts.
\begin{definition}[Graph-based Spatial Model]
The graph-based spatial model with a hierarchy structure $\mathcal{G}=(\mathcal{V},\mathcal{E})$  is constructed by the following rules.
1) The node set $\mathcal{V}=\{V_1,...,V_n\}$ is consisted of a group of node set where each node set $V_k$ represents a finite subset spatial objects from $\Omega_i$. Denote the number of nodes for node set $V_k$ as $n_k$. At each layer, $V_k=\{v_{k,1},...,v_{k,n_k}\}$ contains nodes which represent $n_k$ spatial objects in $\Omega_i$.
2) The edge set $\mathcal{E}$ is used to model the relationship between nodes, such as whether two nodes are adjacent or if one node is included within another node. $e_{i,j}\in \mathcal{E}$ if and only if $v_i$ and $v_j$ are connected. 
3) $v_{k,i}$ is a \emph{parent} of $v_{k+1,j}$, $\forall k\in[1,...,n-1]$, if and only if objects $v_{k+1,i}$ belongs to objects $v_{k,i}$. $v_{k+1,j}$ is called a \emph{child} of $v_{k,i}$ if $v_{k,i}$ is its parent. Furthermore, if $v_i$ and $v_j$ are a pair of parent-child, then $e_{i,j}\in\mathcal{E}$. $v_i$ is a \emph{neighbor} of $v_j$ and $e_{i,j}\in\mathcal{E}$ if and only if there exist $k$ such that $v_i\in V_k$, $v_j\in V_k$, and the minimal distance between $v_i$ and $v_j$ is less than a given threshold $\epsilon$. 
\end{definition}
An example is given in Fig. \ref{exp:Graph with a hierarchy structure} to illustrate the proposed spatial model. In Fig. \ref{exp:Graph with a hierarchy structure}, $V_1=\{kitchen\}$ , $V_2=\{body~part$, $tool$, $material\}$ and $V_3=\{head$, $hand$, $cup$, $bowl$, $table$, $milk$, $butter\}$. The parent-child relationships are drawn in solid lines, and the neighbor relationships are drawn in dashed lines. Each layer represents the space with different spatial concepts or objects by taking categorical values from $\Omega_i$, and connections are built between layers. The hierarchical graph can express facts such as ``head is part of a body part" and ``cup holds milk."

Based on the temporal and spatial model above, the spatial temporal signals we are interested in are defined as follow.
\begin{definition}[Spatial Temporal Signal]
A spatial temporal signal $x(v,t)$ is defined as a scalar function for node $v$ at time $t$
\begin{equation}
    x(v,t): V\times T \rightarrow D,
\end{equation}
where $D$ is the signal domain.
\end{definition}

\subsection{Graph-based spatial temporal logic}
With the temporal model and spatial model in mind, we now give the formal syntax and semantics definition of GSTL. 

\begin{definition}[GSTL Syntax]\label{definition: GSTL syntax}
The syntax of a GSTL formula is defined recursively as 
\begin{equation}
    \begin{aligned}
    &\tau := \mu ~|~ \neg\tau ~|~ \tau_1\wedge\tau_2 ~|~ \tau_1\vee\tau_2 ~|~ \mathbf{P}_A\tau ~|~ \mathbf{C}_A \tau ~|~ \mathbf{N}_A^{\left \langle *, *, * \right \rangle} \tau,\\
    &\varphi := \tau ~|~ \neg \varphi ~|~ \varphi_1\wedge\varphi_2 ~|~ \varphi_1\vee\varphi_2 ~|~ \Box_{[\alpha,\beta]} \varphi~| ~ \varphi_1\sqcup_{[\alpha,\beta]}^{*}\varphi_2.
    \end{aligned}
    \label{complexity proof 1}
\end{equation}
where $\tau$ is spatial term and $\varphi$ is the GSTL formula; $\mu$ is an atomic predicate (AP), negation $\neg$, conjunction $\wedge$ and disjunction $\vee$ are the standard Boolean operators;  $\Box_{[\alpha,\beta]}$ is the ``always" operator and $\sqcup_{[\alpha,\beta]}^{*}$ is the ``until" temporal operators with an Allen interval algebra extension, where $[\alpha,\beta]$ being a real positive closed interval and $*\in\{b, o, d, \equiv, m, s, f\}$ is one of the seven temporal relationships defined in the Allen interval algebra. Spatial operators are ``parent" $\mathbf{P}_A$, ``child" $\mathbf{C}_A$, and ``neighbor" $\mathbf{N}_A^{\left \langle *, *, * \right \rangle}$, where $A$ denotes the set of nodes which they operate on. Same as the until operator, $*\in\{b, o, d, \equiv, m, s, f\}$. 
\end{definition}

The parent operator $\mathbf{P}_A$ describes the behavior of the parent of the current node. The child operator $\mathbf{C}_A$ describes the behavior of children of the current node in the set $A$. The neighbor operator $\mathbf{N}_A^{\left \langle *, *, * \right \rangle}$ describes the behavior of neighbors of the current node in the set $A$. 

We first define an interpretation function before the semantics definition of GSTL. The interpretation function $\iota(\mu,x(v,t)): AP\times D \rightarrow R$ interprets the spatial temporal signal as a number based on the given atomic proposition $\mu$. The qualitative semantics of the GSTL formula is given as follows.

\begin{definition}[GSTL Qualitative Semantics]
The satisfiability of a GSTL formula $\varphi$ with respect to a spatial temporal signal $x(v,t)$ at time $t$ and node $v$ is defined inductively as follows.
\begin{enumerate}
\item $x(v,t)\models \mu$, if and only if $\iota(\mu,x(v,t))>0$;
\item $x(v,t)\models \neg\varphi$, if and only if $\neg(x(v,t))\models \varphi)$;
\item $x(v,t)\models \varphi\land\psi$, if and only if $x(v,t)\models \varphi$ and $x(v,t)\models \psi$;
\item $x(v,t)\models \varphi\lor\psi$, if and only if $x(v,t)\models \varphi$ or $x(v,t)\models \psi$;
\item $x(v,t)\models \Box_{[\alpha,\beta]}\varphi$, if and only if $\forall t'\in[t+\alpha,t+\beta]$, $x(v,t')\models \varphi$;
\item $x(v,t)\models \Diamond_{[\alpha,\beta]}\varphi$, if and only if $\exists t'\in[t+\alpha,t+\beta]$, $x(v,t')\models \varphi$;
%\item $x(v,t)\models \varphi\sqcup_{[a,b]}\psi$, if and only if $\exists t_{k'}\in[t_k+a,t_k+b]$ such that $x(v,t_{k'})\models \psi$ and $\forall t_{k''}\in[t_k,t_{k'}]$, $x(v,t_{k''})\models \varphi$;
\end{enumerate}
The until operator with interval algebra extension is defined as follow.
\begin{enumerate}
\item $({\bf x},t_k)\models \varphi\sqcup_{[\alpha,\beta]}^b\psi$, if and only if $({\bf x},t_k)\models\Box_{[\alpha,\beta]}\neg(\varphi\vee\psi)$ and $\exists t_1<\alpha,~\exists t_2>\beta$ such that $({\bf x},t_k)\models\Box_{[t_1,\alpha]}(\varphi\wedge\neg\psi)\wedge\Box_{[\beta,t_2]}(\neg\varphi\wedge\psi)$;
\item $({\bf x},t_k)\models \varphi\sqcup_{[\alpha,\beta]}^o\psi$, if and only if $({\bf x},t_k)\models\Box_{[\alpha,\beta]}(\varphi\wedge\psi)$ and $\exists t_1<\alpha,~\exists t_2>\beta$ such that $({\bf x},t_k)\models\Box_{[t_1,\alpha]}(\varphi\wedge\neg\psi)\wedge\Box_{[\beta,t_2]}(\neg\varphi\wedge\psi)$;
\item $({\bf x},t_k)\models \varphi\sqcup_{[\alpha,\beta]}^d\psi$, if and only if $({\bf x},t_k)\models\Box_{[\alpha,\beta]}(\varphi\wedge\psi)$ and $\exists t_1<\alpha,~\exists t_2>\beta$ such that $({\bf x},t_k)\models\Box_{[t_1,\alpha]}(\neg\varphi\wedge\psi)\wedge\Box_{[\beta,t_2]}(\neg\varphi\wedge\psi)$;
\item $({\bf x},t_k)\models \varphi\sqcup_{[\alpha,\beta]}^\equiv\psi$, if and only if $({\bf x},t_k)\models\Box_{[\alpha,\beta]}(\varphi\wedge\psi)$ and $\exists t_1<\alpha,~\exists t_2>\beta$ such that $({\bf x},t_k)\models\Box_{[t_1,\alpha]}(\neg\varphi\wedge\neg\psi)\wedge\Box_{[\beta,t_2]}(\neg\varphi\wedge\neg\psi)$;
\item $({\bf x},t_k)\models \varphi\sqcup^m\psi$, if and only if $\exists t_1<t<t_2$ such that $({\bf x},t_k)\models\Box_{[t_1,t]}(\varphi\wedge\neg\psi)\wedge\Box_{[t,t_2]}(\neg\varphi\wedge\psi)$;
\item $({\bf x},t_k)\models \varphi\sqcup^s\psi$, if and only if $\exists t_1<\alpha<\beta<t_2$ such that $({\bf x},t_k)\models\Box_{[t_1,\alpha]}(\neg\varphi\wedge\neg\psi)\wedge\Box_{[\alpha,\beta]}(\varphi\wedge\psi)\wedge\Box_{[\beta,t_2]}(\neg\varphi\wedge\psi)$;
\item $({\bf x},t_k)\models \varphi\sqcup^f\psi$, if and only if $\exists t_1<\alpha<\beta<t_2$ such that $({\bf x},t_k)\models\Box_{[t_1,\alpha]}(\neg\varphi\wedge\psi)\wedge\Box_{[\alpha,\beta]}(\varphi\wedge\psi)\wedge\Box_{[\beta,t_2]}(\neg\varphi\wedge\neg\psi)$;
\end{enumerate}
The spatial operators are defined as follows.
\begin{enumerate}
\item $x(v,t)\models \mathbf{P}_A\tau$, if and only if $\forall v_p\in A,~x(v_p,t)\models \tau$ where $v_p$ is the parent of $v$;
\item $x(v,t)\models \mathbf{C}_{A}\tau$, if and only if $\forall v_c\in A,~x(v_c,t)\models \tau$ where $v_c$ is a child of $v$;
\item $x(v,t)\models \mathbf{N}_{A}^{\left \langle b, *, * \right \rangle}\tau$, if and only if $\forall v_n \in A,~x(v_n,t)\models \tau$ where $v_n$ is a neighbor of $v$ and $v_n[x^+]<v[x^-]$;
\item $x(v,t)\models \mathbf{N}_{A}^{\left \langle o, *, * \right \rangle}\tau$, if and only if $\forall v_n \in A,~x(v_n,t)\models \tau$ where $v_n$ is a neighbor of $v$ and $v_n[x^-]<v[x^-]<v_n[x^+]<v[x^+]$;
\item $x(v,t)\models \mathbf{N}_{A}^{\left \langle d, *, * \right \rangle}\tau$, if and only if $\forall v_n \in A,~x(v_n,t)\models \tau$ where $v_n$ is a neighbor of $v$ and $v_n[x^-]<v[x^-]<v[x^+]<v_n[x^+]$;
\item $x(v,t)\models \mathbf{N}_{A}^{\left \langle \equiv, *, * \right \rangle}\tau$, if and only if $\forall v_n \in A,~x(v_n,t)\models \tau$ where $v_n$ is a neighbor of $v$ and $v_n[x^-]=v[x^-],~v[x^+]=v_n[x^+]$;
\item $x(v,t)\models \mathbf{N}_{A}^{\left \langle m, *, * \right \rangle}\tau$, if and only if $\forall v_n \in A,~x(v_n,t)\models \tau$ where $v_n$ is a neighbor of $v$ and $v_n[x^+]=v[x^-]$;
\item $x(v,t)\models \mathbf{N}_{A}^{\left \langle s, *, * \right \rangle}\tau$, if and only if $\forall v_n \in A,~x(v_n,t)\models \tau$ where $v_n$ is a neighbor of $v$ and $v_n[x^-]=v[x^-]$, $v_n[x^+]>v[x^+]$;
\item $x(v,t)\models \mathbf{N}_{A}^{\left \langle f, *, * \right \rangle}\tau$, if and only if $\forall v_n \in A,~x(v_n,t)\models \tau$ where $v_n$ is a neighbor of $v$ and $v_n[x^+]=v[x^+]$, $v_n[x^-]<v[x^-]$.
\end{enumerate}
\end{definition}
Here $v[x^-]$ and $v[x^+]$ denote the lower and upper limit of node $v$ in x-direction. Definition for the neighbor operator in y-direction and z-direction is omitted for simplicity. Notice that the reverse relations in IA can be easily defined by changing the order of the two GSTL formulas involved, e.g., $\varphi\sqcup_{[\alpha,\beta]}^{o^{-1}}\psi\Leftrightarrow\psi\sqcup_{[\alpha,\beta]}^o\varphi$. 
%As usual, $\varphi_1\vee\varphi_2$, $\varphi_1\rightarrow\varphi_2$, $\varphi_1\leftrightarrow\varphi_2$ abbreviate $\neg(\neg\varphi_1\wedge\varphi_2)$, $\neg\varphi_1\vee\varphi_2$, $(\varphi_1\rightarrow\varphi_2)\wedge(\varphi_2\rightarrow\varphi_1)$, respectively.
Based on the IA relations, we can define six spatial directions (e.g. left, right, front, back, top, down) for the ``neighbor" operator. For instance, $\mathbf{N}_A^{left}=\mathbf{N}_A^{\left \langle *,+,+\right \rangle}$, where $*\in\{b,m\}$ and $+\in\{d,\equiv,o\}$.
We further define another six spatial operators $\mathbf{P}_{\exists}\tau$, $\mathbf{P}_{\forall}\tau$, $\mathbf{C}_{\exists}\tau$, $\mathbf{C}_{\forall}\tau$, $\mathbf{N}_{\exists}^{\left \langle *, *, * \right \rangle}\tau$ and $\mathbf{N}_{\forall}^{\left \langle *, *, * \right \rangle}\tau$ based on the definition above. 
\begin{align*}
    &\mathbf{P}_{\exists}\tau=\vee_{i=1}^{n_p}\mathbf{P}_{A_i}\tau,
    ~\mathbf{P}_{\forall}\tau=\wedge_{i=1}^{n_p}\mathbf{P}_{A_i}\tau,~A_i=\{v_{p,i}\},\\
    &\mathbf{C}_{\exists}\tau=\vee_{i=1}^{n_c}\mathbf{C}_{A_i}\tau,
    ~\mathbf{C}_{\forall}\tau=\wedge_{i=1}^{n_c}\mathbf{C}_{A_i}\tau,~A_i=\{v_{c,i}\},\\
    %&\mathbf{C}_{\forall}\varphi=\vee_{i=1}^{n_c}\mathbf{C}_{A_i}\varphi,\\
    &\mathbf{N}_{\exists}^{\left \langle *, *, * \right \rangle}\tau=\vee_{i=1}^{n_n}\mathbf{N}_{A_i}^{\left \langle *, *, * \right \rangle}\tau,
    ~\mathbf{N}_{\forall}^{\left \langle *, *, * \right \rangle}\tau=\wedge_{i=1}^{n_n}\mathbf{N}_{A_i}^{\left \langle *, *, * \right \rangle}\tau,~A_i=\{v_{n,i}\},
\end{align*}
where $v_{p,i}$, $v_{c,i}$, $v_{n,i}$ are the parent, child, and neighbor of $v$ respectively and $n_c$, $n_n$ are the number of children and neighbors of $v$ respectively.

As we can see from the syntax definition of GSTL, the definition implies the following assumption, which is reasonable to applications such as autonomous robots.
\begin{assumption}[Domain closure]
The only objects in the domain are those representable using the existing symbols, which do not change over time.
\end{assumption}
The restriction that no temporal operators are allowed in the spatial term is reasonable for robotics since usually predicates are used to represent objects such as cups and bowls. We do not expect cups to change to bowls over time. Thus, we do not need any temporal operator in the spatial term and adopt the following assumption.

\section{Specification mining based on video}\label{section:specification mining}

One of the key steps in employing spatial temporal logics for autonomous robots is specification mining. Specifically, it is crucial for autonomous robots to learn new information in the form of GSTL formulas from the environment via sensor (e.g., video) directly without human inputs. In this section, we introduce an algorithm of mining a set of parametric GSTL formulas through specification mining based on a demo video.

We first pre-processed the video and stored each video as a sequence of graphs: $\mathcal{G}^i=(G_{1}^i,...,G_{T}^i)$, where $G_{t}^i=(V_{t}^i,W_{t}^i)$. $G_{t}^i$ represents frame $t$ in the original video $i$. $V_{t}^i=(v_{t,1}^i,...,v_{t,k}^i)$ stores objects in frame $t$ where $v_{t,k}^i$ is the object such as ``cup" and ``hand". $w_{t,i,j}^i\in W_t^i$ stores the 3D directional information (e.g. left, right, front, back, top, down) between object $v_{t,i}^i$ and object $v_{t,j}^i$ at frame $t$. $w_{t,i,j}^i$ can be obtained easily based on the 3D information returned by the stereo camera.

The specification mining procedure is introduced as follows, with an example to illustrate the algorithm. The basic idea of the proposed specification mining is to first build spatial terms inductively and then construct more complicated temporal formulas by assembling the spatial terms from the previous steps. Specifically, for each frame $G_t^i$ of the video $i$, we generate spatial terms $\nu$ for each objects detected in $G_t^i$. For autonomous robots, both connectivity and parthood spatial relations are crucial to make decisions. Thus, we need to mine both of them from each frames. 

For connectivity, we generate spatial terms $\nu_1\wedge\mathbf{N}_\exists^{\left \langle *, *, * \right \rangle}\nu_2$ if the distance of the objects represented by spatial terms $\nu_1$ and $\nu_2$ is less than a given threshold $d$. The direction variable $*$ is replaced with proper 3D spatial relations between the objects represented by $\nu_1$ and $\nu_2$. If the relative position of the two objects satisfies at any six directions defined in the neighbor operator semantic definition, we replace ${\left \langle *, *, * \right \rangle}$ with corresponding directional relations. For parthood, we generate spatial terms $\mathbf{P}_\exists\nu$ for each object in $G_t^i$. As we mentioned in Section \ref{problem formulation and assumption}, we assume we know parthood relations for objects we are interested in.  

Next, we build more complicated spatial terms by combining spatial terms from the previous steps. Let us assume the hierarchical graph defined in GSTL has three layers. We obtain the following spatial terms for frame $G_t^i$ which include both connectivity and parthood spatial information.
\begin{equation}
    \begin{aligned}
    &\mathbf{C}_\exists^2(\nu_1\wedge\mathbf{N}_\exists^{\left \langle *, *, * \right \rangle}\nu_2),~
    \mathbf{C}_\exists^2(\mathbf{P}_\exists\nu),\\
    &\mathbf{C}_\exists^2(\nu_1\wedge\mathbf{N}_\exists^{\left \langle *, *, * \right \rangle}\mathbf{P}_\exists\nu_2),~
    \mathbf{C}_\exists^2(\mathbf{P}_\exists\nu_1\wedge\mathbf{N}_\exists^{\left \langle *, *, * \right \rangle}\mathbf{P}_\exists\nu_2).
\end{aligned}
\label{spatial terms templates}
\end{equation}
For an example in Fig \ref{hsvfilter}, we can generate the following GSTL terms for $G_t^i$, where $r_1$ states that the cup is behind the plate, $r_2$ states that the fork is at the left of the plate, $r_3$ states that the hand is grabbing a cup. $r_4$ states there are tools in the current frame. $r_5$ states hand is operating tools. $r_6$ states body parts and tools are connected in the current frame. Some GSTL terms are omitted for simplicity. 
\begin{align*}
    &r_1=\mathbf{C}_\exists^2(cup\wedge\mathbf{N}_\exists^{behind} plate),\\
    &r_2=\mathbf{C}_\exists^2(fork\wedge\mathbf{N}_\exists^{right} plate),\\
    &r_3=\mathbf{C}_\exists^2(hand\wedge\mathbf{N}_\exists^{\left \langle b, b, b^{-1} \right \rangle} cup),\\
    &r_4=\mathbf{C}_\exists^2(\mathbf{P}_\exists tool),~
    r_5=\mathbf{C}_\exists^2(hand\wedge\mathbf{N}_\exists^{\left \langle b, b, b^{-1} \right \rangle}\mathbf{P}_\exists tool),\\
    &r_6=\mathbf{C}_\exists^2(\mathbf{P}_\exists body ~part\wedge\mathbf{N}_\exists^{\left \langle b, b, b^{-1} \right \rangle}\mathbf{P}_\exists tool).
\end{align*}

To generate temporal formulas, we first merge consecutive $G_t^i$ with the exact same set of GSTL terms from the previous step by only keeping the first and the last frame. For example, assuming for video $i$ from $G_0^i$ to $G_{35}^i$, all frames satisfy $\tau_1$ and $\tau_2$, then we merge them together by only keeping $G_0^i$, $G_{35}^i$, and the GSTL terms they satisfied. Then we generate ``Always" formula based on the frames and terms we kept from the previous step. We find the maximum time interval for each GSTL term from the previous step. For example, $\varphi_1=\Box_{[0,90]}r_1$, $\varphi_2=\Box_{[100,190]}r_2$, and $\psi=\Box_{[45,120]}r_3$.

In theory, the ``Always" GSTL formulas generated by the previous step include all the information from the video. However, it does not show any temporal relations between any two formulas. Thus, we use ``Until" operators to mine more temporal information based on the template with a temporal structure. As our goal is to build a domain theory focusing on available actions, we are interested in what can ``hand" do to other tools. Specifically, we want to generate a motion primitive in GSTL formulas which includes the action itself and the preconditions and effects of the action. Thus, we generate the following GSTL formula 
\begin{equation}
    \begin{aligned}
    &a=(\Box_{[t_1,t_2]}\tau_1)\sqcup_{[\alpha_1,\beta_1]}^o(\Box_{[\alpha,\beta]}\tau_2)\sqcup_{[\alpha_2,\beta_2]}^o(\Box_{[t_3,t_4]}\tau_3),\\
    &\tau_1=\mathbf{C}^2_{\exists}(\nu_1\wedge\mathbf{N}_{\exists}^{\left \langle *, *, * \right \rangle}\nu_2),~\nu_1\models\mathbf{P}_\exists tool,~\nu_2\models\mathbf{P}_\exists tool,\\
    &\tau_2=\mathbf{C}^2_{\exists}(hand\wedge\mathbf{N}_{\exists}^{\left \langle *, *, * \right \rangle}\nu_1),~\nu_1\models\mathbf{P}_\exists tool,\\
    &\tau_3=\mathbf{C}^2_{\exists}(\nu_1\wedge\mathbf{N}_{\exists}^{\left \langle *, *, * \right \rangle}\nu_3),~\nu_3\models\mathbf{P}_\exists tool.
\end{aligned}
\label{templates}
\end{equation}
As we can see from \eqref{templates}, $\tau_1$, $\tau_2$, and $\tau_3$ share the same object $\nu_1$ because the hand is operating the object. We check if any ``Always" GSTL formulas satisfy \eqref{templates}. If so, we generate a GSTL formula by replacing $\tau_i$ with the ``Always" GSTL formulas. Let us continue the previous example. Denote the ``Always" formulas generated from the previous step as $\varphi_i$ and $\varphi_j$ ( formulas without hands) and $\psi_i$ (formulas with hands). If $\varphi_i$ and $\varphi_j$ have common objects and $\psi_i$ operates the object, then we check if they have the temporal relationship defined in \eqref{templates} if and only if their time intervals satisfy the \emph{overlap} relations defined in Allen's interval algebra. For example, $\varphi_1$, $\varphi_2$, and $\psi$ satisfy $a_1=\varphi_1\sqcup_{[45,90]}^{o}\psi\sqcup_{[100,120]}^{o}\varphi_2$. 

In the end, we replace the time stamps in the formulas with temporal variables as specific time instances do not apply to other applications. The specification mining algorithm is summarized in Algorithm \ref{specification mining algorithm}. The proposed specification mining based on video algorithm terminates in finite time for finite length video since the number of GSTL terms and formulas one can get is finite.
\begin{algorithm}[ht]
\SetAlgoLined
\LinesNumbered
\SetKwInOut{Input}{input}\SetKwInOut{Output}{output}
\Input{A set of video and parametric GSTL formulas templates $a$ in \eqref{templates}}
\Output{Parametric GSTL formulas}
\BlankLine
 For each video $i$, pre-process the video and stored each video as a sequence of graph $\mathcal{G}^i=(G_1^i,...,G_T^i)$\;
 \For{for each $G_t^i$}{
 \For{for object $v_{t,k}^i$ in $G_t^i$ }{
 Generate $\mathbf{C}_\exists^2(\mathbf{P}_\exists v_{t,k}^i)$ using the parthood information of $v_{t,k}^i$\;
 \If{the distance between $v_{t,k}^i$ and $v_{t,l}^i$ in $G_t^i$ is smaller than a given distance $d$}{
 \If{$v_{t,k}^i$ and $v_{t,l}^i$ satisfy any formulas in \eqref{spatial terms templates}}{
 Replace the $\nu_1$ and $\nu_2$ with $v_{t,k}^i$ and $v_{t,l}^i$\;
 Replace the directional variable $*$ with the corresponding 3D IA directions\;
 }
 }
 }
 }
 Merge consecutive $G_t^i$ with the exact same set of GSTL terms by only keeping the first and the last frame and generate ``Always" formula based on the frame and terms\;
 \If{``Always" GSTL formulas satisfy GSTL formula $a$ in \eqref{templates}}{
 Replace $\tau_i$ with the ``Always" GSTL formulas\;
 Output the parametric GSTL formula\;
 }
 
 Output the parametric GSTL formula.
 \caption{Specification mining based on demo videos}
 \label{specification mining algorithm}
\end{algorithm}

\section{Automatic task planner based on domain theory}\label{section:task planning}
From the specification mining through video, robots are able to learn a set of available actions to alter the environment along with its preconditions and effects.
In this section, we focus on developing a task planner for autonomous robots to generate a detailed task plan from a vague task assignment using the available actions learned from the previous section. We first introduce the domain theory, which stores necessary information to accomplish the task for autonomous robots. Then we introduce the automatic task planner composed of the proposer and the verifier. In the end, an overall framework for autonomous robots that combines the task planner and domain theory is given. 

\subsection{Domain theory}
On the one hand, the proposed GSTL formulas can be used to represent knowledge for autonomous robots. On the other hand, robots need a set of knowledge or common sense to solve a new task assignment. Thus, we define a domain theory in GSTL  for autonomous robots which stores available actions for robots to solve a new task. In the domain theory, the temporal parameters are not fixed. The domain theory is defined as a set of parametric GSTL formulas as follows.
\begin{definition}
Domain theory $\Sigma$ is a set of parametric GSTL formulas that satisfies the following consistent condition.
\begin{itemize}
    \item Consistent: $\forall \varphi_i,\varphi_j\in \Sigma$, there exists a set of parameters such as $\varphi_i\wedge\varphi_j$ is true.
    %\item Minimum: $\forall \varphi_i\in\Sigma$, there does not exist a set of parameters such as $\Sigma\setminus \{\varphi_i\}\Rightarrow\varphi_i$ is true.
\end{itemize}
\end{definition}
For example, the set $\Sigma$ including the following parametric GSTL formulas is a domain theory.
\begin{equation}
            \begin{aligned}
            &\Box_{[t_1,t_2]}\mathbf{C}_\exists (tools \wedge \mathbf{C}_\exists (cup\vee plate \vee fork \vee spoon)) \\
            %&\Box_{[t_1,t_2]}\mathbf{C}_\exists (cupboard \wedge \mathbf{C}_\exists (flour \vee baking powder \vee salt \vee sugar)) \\
            %&\Box_{[t_1,t_2]}\mathbf{C}_\exists^2 (door \wedge \mathbf{N}_\exists^{\left \langle d, d, m \right \rangle} handle),\\
           % &\Box_{[t_1,t_2]}\mathbf{C}_\exists^2 ( \mathbf{P}_\exists utensils \wedge \mathbf{N}_\exists^{\left \langle *, *, * \right \rangle} (table \vee hand \vee plate)),\\
            &\varphi_1= \Box_{[t_1,t_2]}\mathbf{C}_\exists^2 (hand \wedge \mathbf{N}_\exists^{\left \langle *, *, * \right \rangle} cup),\\
            &\varphi_2= \Box_{[t_1,t_2]}\mathbf{C}_\exists^2 (cup \wedge \mathbf{N}_\exists^{\left \langle d, d, m \right \rangle} table),\\
            &\varphi_2 \sqcup_{[\alpha_1,\beta_1]}^o\varphi_1 \sqcup_{[\alpha_2,\beta_2]}^o \varphi_2.
            \end{aligned}
        \end{equation}
It states common sense such that tools includes cup, plate, fork, and spoon and action primitives such as hand grab a cup from a table and put it back after use it. 
\begin{comment}
The following set is not a domain theory as it violates the minimum conditions.
\begin{equation}
            \begin{aligned}
            &\Box_{[t_1,t_2]}\mathbf{C}_\exists (cup \wedge \mathbf{N}_\exists^{\left \langle *, *, * \right \rangle} hand) \\
            &\Box_{[t_1,t_2]}\mathbf{C}_\exists (cup \wedge \mathbf{N}_\exists^{\left \langle *, *, * \right \rangle} (hand\vee table).
            \end{aligned}
\end{equation}
\end{comment}

\begin{remark}[Modular reasoning]
Our domain theory inherits the hierarchical structure from the hierarchical graph from the GSTL spatial model. This is an important feature and can be used to reduce deduction systems complexity significantly. Domain theory for real-world applications often demonstrates a modular-like structure in the sense that the domain theory contains multiple sets of facts with relatively little connection to one another \cite{lifschitz2008knowledge}. For example, a domain theory for kitchen and bathroom will include two sets of relatively self-contained facts with a few connections such as tap and switch. A deduction system that takes advantage of this modularity would be more efficient since it reduces the search space and provides less irrelevant results. Existing work on exploiting the structure of a domain theory for automated reasoning can be found in \cite{amir2005partition}.
\end{remark}

In this paper, we generate the domain theory through specification mining based on the demo video. Using the algorithm from the previous section, we have the following domain theory, which will be presented in the evaluation section. The domain theory in \eqref{knowledge base} is used in automated task planning. Notice that the domain theory does not limit to any specific initial state. The domain theory can be applied to any table set up involving cup, plate, spoon, and fork.

\begin{equation}
    \begin{aligned}
    &s_1=\mathbf{C}^2_{\exists}(cup\wedge\mathbf{N}_{\exists}^{back}plate),~
    s_2=\mathbf{C}^2_{\exists}(fork\wedge\mathbf{N}_{\exists}^{left}cup),~\\
    &s_3=\mathbf{C}^2_{\exists}(spoon\wedge\mathbf{N}_{\exists}^{left}fork),
    s_4=\mathbf{C}^2_{\exists}(fork\wedge\mathbf{N}_{\exists}^{left}empty),\\
    &s^*_1=\mathbf{C}^2_{\exists}(cup\wedge\mathbf{N}_{\exists}^{top}plate),~
    s^*_2=\mathbf{C}^2_{\exists}(fork\wedge\mathbf{N}_{\exists}^{left}plate),~\\
    &s^*_3=\mathbf{C}^2_{\exists}(spoon\wedge\mathbf{N}_{\exists}^{right}plate),\\
&a_1^{'}=\mathbf{C}_\exists^2(hand\wedge\mathbf{N}_\exists^{{\left \langle *, *, * \right \rangle}}cup),~
a_2^{'}=\mathbf{C}_\exists^2(hand\wedge\mathbf{N}_\exists^{{\left \langle *, *, * \right \rangle}}fork),\\ &a_3^{'}=\mathbf{C}_\exists^2(hand\wedge\mathbf{N}_\exists^{{\left \langle *, *, * \right \rangle}}spoon),~ a_4^{'}=\mathbf{C}_\exists^2(hand\wedge\mathbf{N}_\exists^{{\left \langle *, *, * \right \rangle}}plate),\\
    &a_1=(\Box_{[t_1,t_2]}s_1)\sqcup_{[\alpha_1,\beta_1]}^o (\Box_{[t_3,t_4]} a_1^{'}) \sqcup_{[\alpha_2,\beta_2]}^o(\Box_{[t_5,t_6]}s^*_1),~\\
    &a_2=(\Box_{[t_1,t_2]}s_2)\sqcup_{[\alpha_1,\beta_1]}^o (\Box_{[t_3,t_4]} a_2^{'}) \sqcup_{[\alpha_2,\beta_2]}^o(\Box_{[t_5,t_6]}s^*_2),~\\
    &a_3=(\Box_{[t_1,t_2]}s_3)\sqcup_{[\alpha_1,\beta_1]}^o (\Box_{[t_3,t_4]} a_3^{'})\sqcup_{[\alpha_2,\beta_2]}^o(\Box_{[t_5,t_6]}s^*_3),~\\
    &a_4=(\Box_{[t_1,t_2]}s_4)\sqcup_{[\alpha_1,\beta_1]}^o (\Box_{[t_3,t_4]} a_4^{'})\sqcup_{[\alpha_2,\beta_2]}^o(\Box_{[t_5,t_6]}s^*_2).
\end{aligned}
\label{knowledge base}
\end{equation}

\subsection{Control synthesis}
The task planner takes environment information from sensors and available actions from the domain theory and solves a vague task assignment with detailed task plans. For example, we give a task assignment to a robot by asking it to set up a dining table. A camera will provide the current dining table setup, and the domain theory stores information on what actions robots can take. The goal for the task planner is to generate a sequence of actions robots need to take such that the robot can set up the dining table as required. Specifically, we propose to implement the task planner as two interacting components, namely proposer and verifier. The proposer first proposes a plan based on the domain theory and its situational awareness. The verifier then checks the feasibility of the proposed plan based on the domain theory. If the plan is not feasible, then it will ask the proposer for another plan. 
%For example, if the initial plans generated by the proposer includes ``hand touch hot water" while the domain theory specifies constraints ``hand cannot touch hot materials," then the verifier will find the conflicts and inform the proposer that ``hand cannot touch hot water" needed to be considered in the re-planning. The proposer may come up with a new plan where the hand will use a cup to hold hot water. 
If the plan turns out to be feasible, the verifier will output the plan to the robot for execution. The task planner may be recalled once the situation changes during the execution.  

\subsubsection{Proposer}
For the proposer, we are inspired to cast the planning as a path planning problem on a graph $M=(\mathcal{S}, \mathcal{A},T)$ as shown in Fig. \ref{exp:task planning}, where node $s_i\in \mathcal{S}$ represents a GSTL term for objects that hold true at the current status. The initial term $s_0$ corresponding to a point or a set of points in $\mathcal{S}$, while the target terms $s^*_1$, $s^*_2$, and $s^*_3$ in $\mathcal{S}$ corresponds to the accomplishment of the task. Actions $a$ available to robots are given in $\mathcal{A}$ as GSTL formulas from the domain theory. The transition function $T: \mathcal{S}\times \mathcal{A} \rightarrow \mathcal{S}$ mapping one spatial term to another is triggered by an action $a\in \mathcal{A}$ that the robot can take. The example is given in Fig. \ref{exp:task planning example} where the goal is to set up the dinner table as shown in the right figure. The initial states $s_1,s_2,s_3$, target states $s^*_1,s^*_2,s^*_3$, and available actions $a_1,a_2,a_3,a_4$ are given in the domain theory (\ref{knowledge base}).

The goal of the proposer is to find an ordered set of actions that transform initial spatial terms into target terms. It is worth pointing out such a graph in Fig. \ref{exp:task planning} is not given \emph{a prior} to robots. Robots need to expand the graph and generate serial actions by utilizing information in the domain theory. Similar to the task planning in GRAPHPLAN \cite{weld1999recent}, the proposer generates a potential solution in two steps, namely forward graph expansion and backward solution extraction.  

In the graph expansion, we expand the graph forward in time until the current spatial terms level includes all target terms, and none of them is mutually exclusive. To expand the graph, we start with initial terms and expanding the graph by applying available actions to the terms. The resulting terms based on the transition function $T$ will be new current terms. We define an exclusive mutual relation (mutex) for actions and terms and label all mutex relations among actions and terms. Two actions are mutex if they satisfy one of the following conditions. 1) The effect of one action is the negation of the other action. 2) The effect of one action is the negation of the other action's precondition. 3) Their preconditions are mutex. Furthermore, we say two terms are mutex if their supporting actions are mutex. If the current term level includes all target terms, and there is no mutex among them, then we move to the solution extraction phase as a solution may exist in the current transition system. The algorithm is summarized in Algorithm \ref{algorithm:proposer:expansion}, where $\mathcal{A}_{s_i^k}$ is the set of supporting actions for $s_{i}^{k}$.

\begin{algorithm}[ht]
\SetAlgoLined
\LinesNumbered
\SetKwInOut{Input}{input}\SetKwInOut{Output}{output}
\Input{Observed terms $s_1,s_2,...,s_n$, available actions $a_1,a_2,...,a_l$, target term $s^*_1,s^*_2,...,s^*_m$;}
\Output{A graph with all target terms included;}
\BlankLine

 Initialization: $s^0=\{s_1^0,s_2^0,...,s_n^0\}$ and $s^*=\{s^*_1,s^*_2,...,s^*_m\}$\;
 \While{$\exists s^*_i\not\in s^k$ or $\exists s^*_i ~\text{and}~ s^*_j \in s^k$ which are mutex}{
  For all $s_i^k$ at current level $k$, add $s_j^{k+1}$ into the next level $k+1$ if $s_j^{k+1}\in \{s_i^k\times a_i\}$\;
  \If{effect of $a_i$ is the negation of the precondition of $a_j$ or $a_i$ and $a_j$ have conflict preconditions or the effects of $a_i$ and $a_j$ are mutex}{
   Add a mutex link between $a_i$ and $a_j$\;
   }
 
 \If{$\forall a_i\in\mathcal{A}_{s_{i}^{k+1}}$ and $\forall a_j\in\mathcal{A}_{s_{j}^{k+1}}$, $a_i$ and $a_j$ are mutex}{Add a mutex link between $s_i^{k+1}$ and $s_j^{k+1}$\;}
}
 \caption{Task planning for the graph expansion phase of the proposer}
 \label{algorithm:proposer:expansion}
\end{algorithm}

In the solution extraction phase, we extract solution backward by starting with the current term level. For each target term ${s^*_i}^k$ at the current term level $k$, we denote the set of its supporting actions as $\mathcal{A}_{{s^*_i}^{k}}$. We choose one action from each $\mathcal{A}_{{s^*_i}^{k}}$ with no mutex relations for all target terms, formulate a candidate solution set at this step $\mathcal{A}^k$, and denote the precondition terms of the selected actions as $\mathcal{S}_{pre}^{k}$. Then we check if the precondition terms have mutex relations. If so, we terminate the search on $\mathcal{A}^k$ and choose another set of action as a candidate solution until we enumerate all possible combinations. If no mutex relations are detected in $\mathcal{S}_{pre}^{k}$, then we repeat the above backtracking step until the mutex is founded or $\mathcal{S}_{pre}^{k}$ includes all initial terms. The solution extraction algorithm is summarized in Algorithm \ref{algorithm:proposer:extraction}. 

\begin{algorithm}[ht]
\SetAlgoLined
\LinesNumbered
\SetKwInOut{Input}{input}\SetKwInOut{Output}{output}
\Input{The transition graph from the expansion phase;}
\Output{A sequence of actions $a_1,a_2,...,a_k$;}
\BlankLine
 $\forall {{s^*_i}}^k\in s^*$ at the current state level $k$, denote the set of its supporting actions as $\mathcal{A}_{{{s^*_i}}^{k}}$\;
 Pick one solution set $\mathcal{A}^k$  by choosing one actions from each set $\mathcal{A}_{{{s^*_i}}^{k}}$ where no mutex relations are allowed\;
  Denote the precondition terms of the selected actions as $\mathcal{S}_{pre}^k$\;
  \eIf{$\mathcal{S}_{pre}^k$ has no mutex terms}{
  \If{$\mathcal{S}_{pre}^k$ includes all initial terms}{Output the ordered actions\;}
   $\forall s_i^k\in \mathcal{S}_{pre}^k$, denote the set of its supporting actions as $\mathcal{A}_{{{s^*_i}}^{k-1}}$\;
 Repeat the extraction process for $\mathcal{A}_{{{s^*_i}}^{k-1}}$ by going to line 2\;
   }
   {Discard the solution $\mathcal{A}^k$ and pick a new solution from $\mathcal{A}_{{{s^*_i}}^k}$ by going to line 2 until all combinations have been enumerated \;
   }
 \If{No feasible solution has been found}{Go to Algorithm \ref{algorithm:proposer:expansion} and expand the graph.}
 \caption{Task planning for the solution extraction phase of the proposer}
 \label{algorithm:proposer:extraction}
\end{algorithm}

\subsubsection{Verifier}
Next, we introduce the implementation of the verifier and the interaction with the proposer.
The verifier checks if the plan generated by the proposer is executable based on constraints in the domain theory and information from the sensors. The plan is executable if the verifier can find a set of parameters for the ordered actions given by the proposer while satisfying all constraints posed by the domain theory and the sensors. If the plan is not executable, then it will ask the proposer for another plan. If the plan is executable, it will output the effective plans to autonomous robots. 

As the task plans from the proposer are parametric GSTL formulas, the verifier needs two steps to verify if the parametric GSTL formulas are feasible. First, the verifier reformulates the parametric GSTL formulas in $\wedge_i(\vee_j\pi_{i,j})$, where $\pi_{i,j}$ is either spatial terms or spatial terms with ``Always" operators. The verifier finds feasible temporal parameters for every terms in $\wedge_i(\vee_j\pi_{i,j})$ using a satisfiability modulo theories (SMT) solver. Then, the verifier checks if there is a feasible solution for the spatial terms by formulating them in CNF and solving it using an SAT solver. We explain the two steps in detail as follows.

We first use SMT to find feasible temporal parameters for the parametric GSTL formulas from the proposer. SMT is the extension of the SAT, where the binary variables are replaced with predicates over a set of non-binary variables. The predicate is a binary function $f(x)\in \mathbb{B}$ with non-binary variables $x\in\mathbb{R}$. The predicate can be interpreted with different theories. For example, the predicate in SMT can be a function of linear inequality, which returns one if and only if the inequality holds true. An example of linear inequality predicate is given below.
\begin{align*}
    (\beta-\alpha<5) \wedge ((\alpha + \beta < 10) \vee (\alpha - \beta > 20)) \wedge (\alpha +\beta >\gamma)
\end{align*}
Here, $\alpha,\beta,\gamma$ are non-binary variables and we use the linear inequality to represent the predicate for simplicity. To obtain the above form $\wedge_i(\vee_j f_{i,j})$ so that we can apply SMT solver, we reformulate the parametric GSTL formulas in the following form,
\begin{equation}
\begin{aligned}
    &\varphi := \wedge(\vee\pi),\\
    & \pi := \tau ~|~ \neg\tau ~|~ \Box_{[t_1,t_2]}\tau ~|~ \Box_{[t_1,t_2]}\neg\tau,\\
    &\tau := \mu ~|~ \neg\tau ~|~ \tau_1\wedge\tau_2 ~|~ \tau_1\vee\tau_2 ~|~ \mathbf{P}_A\tau ~|~ \mathbf{C}_A \tau ~|~ \mathbf{N}_A^{\left \langle *, *, * \right \rangle} \tau.
\end{aligned}
\label{smt form}
\end{equation}
One can write any GSTL formula in $\wedge(\vee \pi)$ because any GSTL formula can be formulated in CNF form as shown in \cite{liu2020graph}. We define the lower temporal bound and the upper temporal bound for each spatial term $\tau$ of $\pi$ in (\ref{smt form}) as two temporal variables (say $\alpha$ and $\beta$) in real domain. If $\tau$ only hold true at current time instance, then $\alpha=\beta$. Then $\pi$ in (\ref{smt form}) can be represented with the following linear inequality predicates.
\begin{equation}
\begin{aligned}
    &\pi=\tau \Leftrightarrow \alpha\leq t\leq \beta,\\
    &\pi=\neg\tau \Leftrightarrow (t< \alpha)\vee(t<\beta),\\
    &\pi=\Box_{[t_1,t_2]}\tau\Leftrightarrow (\alpha\leq t_1) \wedge (t_2\leq \beta),\\
    &\pi=\Box_{[t_1,t_2]}\neg\tau\Leftrightarrow (t_2\geq \alpha)\wedge (t_1\leq \beta),
\end{aligned}
    \label{smt pi}
\end{equation}
where $t$ is the current time. According to the first two lines of (\ref{smt form}) and \eqref{smt pi}, we can formulate any parametric GSTL formulas in the following SMT form.
\begin{equation}
    \begin{aligned}
            &\varphi=\wedge_i(\vee_j f_{i,j}),
    \end{aligned}
    \label{smt}
\end{equation}
where $f_{i,j}$ is the predicates with linear inequality shown in \eqref{smt pi}. We use existing SMT solvers to find a feasible solution for the problem above. If the parametric GSTL formulas are feasible, the solver will return a time interval $[\alpha,\beta]$ for each spatial term $\tau$ in (\ref{smt form}). The spatial term $\tau$ between $[\alpha,\beta]$ must hold true, which will be checked via SAT.

The rest of the verifier is implemented through SAT. Assume we have a set of spatial terms $\Gamma=\{\tau_1,\tau_2,...,\tau_n\}$ whose lower temporal bound $\alpha_i$ and upper temporal bound $\beta_i$ are given by the SMT solver. We aim to check if all spatial terms can hold true in their corresponding time interval. We have shown in \cite{liu2020graph} that any spatial term $\tau$ can be written in CNF $\wedge_i(\vee_j\mu_{i,j})$ by following the Boolean encoding procedure. Then we obtain a set of logic constraints for spatial terms $\mu_{i,j}$ in CNF whose truth value is to be assigned by the SAT solver. This is done by checking the satisfaction of the following formulas.
\begin{equation}
    \begin{aligned}
            &\varphi=\wedge_{k=1}^{n}\tau_i=\wedge_{k=1}^{n}(\wedge_{t=a_k}^{b_k}\tau_{k,t}),\\
            &\tau_{k,t}=\wedge_i(\vee_j\mu_{i,j}^{k,t}).
    \end{aligned}
    \label{sat}
\end{equation}
The solver will give two possible outcomes. First, the solver finds a feasible solution and $\varphi$ holds true. The plans generated by the proposer successfully solve the new task assignment. In this case, the verifier will output the effective plans to robots with temporal parameters generated from SMT solver. Second, the solver cannot find a feasible solution where $\varphi$ holds true which means the task assignment is not accomplished and there are conflicts in the plans generated by the proposer. The verifier will inform the proposer the plan is infeasible. The proposer will take the information as additional constraints and replan the transition system. The algorithm is summarized in Algorithm \ref{algorithm:verfier}.

\begin{algorithm}[ht]
\SetAlgoLined
\LinesNumbered
\SetKwInOut{Input}{input}\SetKwInOut{Output}{output}
\Input{Ordered actions from the proposer $a_1,a_2,...$, task assignment $\psi$, observed event, and domain theory}
\Output{An executable sequence of actions or counter-example}
\BlankLine
 Rewrite ordered action plans as $\phi_i=a_1\sqcup_{[c_1,c_2]}^ba_2\sqcup_{[c_3,c_4]}^b\cdots$ and denote $\Sigma=\{\phi_i,\psi\}$\;
 \While{there are $\Diamond_{[\alpha,\beta]}$ and $\sqcup_{[\alpha,\beta]}^*$ operators in formulas of $\Sigma$}{
 %for GSTL formula $\varphi=\Box_{[a,b]}\phi$, we have $\varphi=\wedge_{i=a}^b\phi_i$ where $\phi_i$ represents $\phi$ at time $i\in[a,b]$\;
 %for GSTL formula $\varphi=\Diamond_{[a,b]}\phi$, we have $\varphi=\vee_{i=a}^b\phi_i$ where $\phi_i$ represents $\phi$ at time $i\in[a,b]$\;
 %for GSTL formula $\varphi_1\sqcup_{[a,b]}^o\varphi_2$, we have $\wedge_{i=a}^{b}(\varphi_{1,i}\wedge\varphi_{2,i})\wedge \varphi_{1,a-1} \wedge \varphi_{2,b+1} \wedge \neg\varphi_{1,b+1} \wedge \neg\varphi_{2,a-1}$ (other IA relations can be transferred in the similar way)\;
 for GSTL formula $\varphi=\Diamond_{[\alpha,\beta]}\phi$, we have $\varphi=\neg\Box_{[\alpha,\beta]}\neg\phi$\;
 for GSTL formula $\varphi_1\sqcup_{[\alpha,\beta]}^o\varphi_2$, we have $\Box_{[\alpha,\beta]}(\varphi_1\wedge\varphi_2)\wedge \Box_{[\alpha-1,\alpha-1]}\neg\varphi_2\wedge\Box_{[\beta-1,\beta-1]}\neg\varphi_1$ (other IA relations can be transferred in the similar way)\;
 }
 Reform $\Sigma$ as (\ref{smt form})\;
 Replace $\pi$ in (\ref{smt form}) with linear inequality predicates based on (\ref{smt pi})\;
 Solve the corresponding SMT in (\ref{smt})\;
 Reform $\tau$ at each time in CNF form (\ref{sat})\;
 Solve the SAT problem in (\ref{sat}) by assigning a set of truth value $u:\tau\rightarrow\{\top,\bot\}\in\mathbf{U}$ to each $\mu_{j,i}^*$\;
 \eIf{a feasible solution has been found}{
   Output the executable ordered action plan\;
   }{
   Inform the proposer such plan is infeasible\;
  }
 \caption{Verifier}
 \label{algorithm:verfier}
\end{algorithm}

\subsection{Overall framework}
\begin{figure}
    \centering
    \includegraphics[scale=0.15]{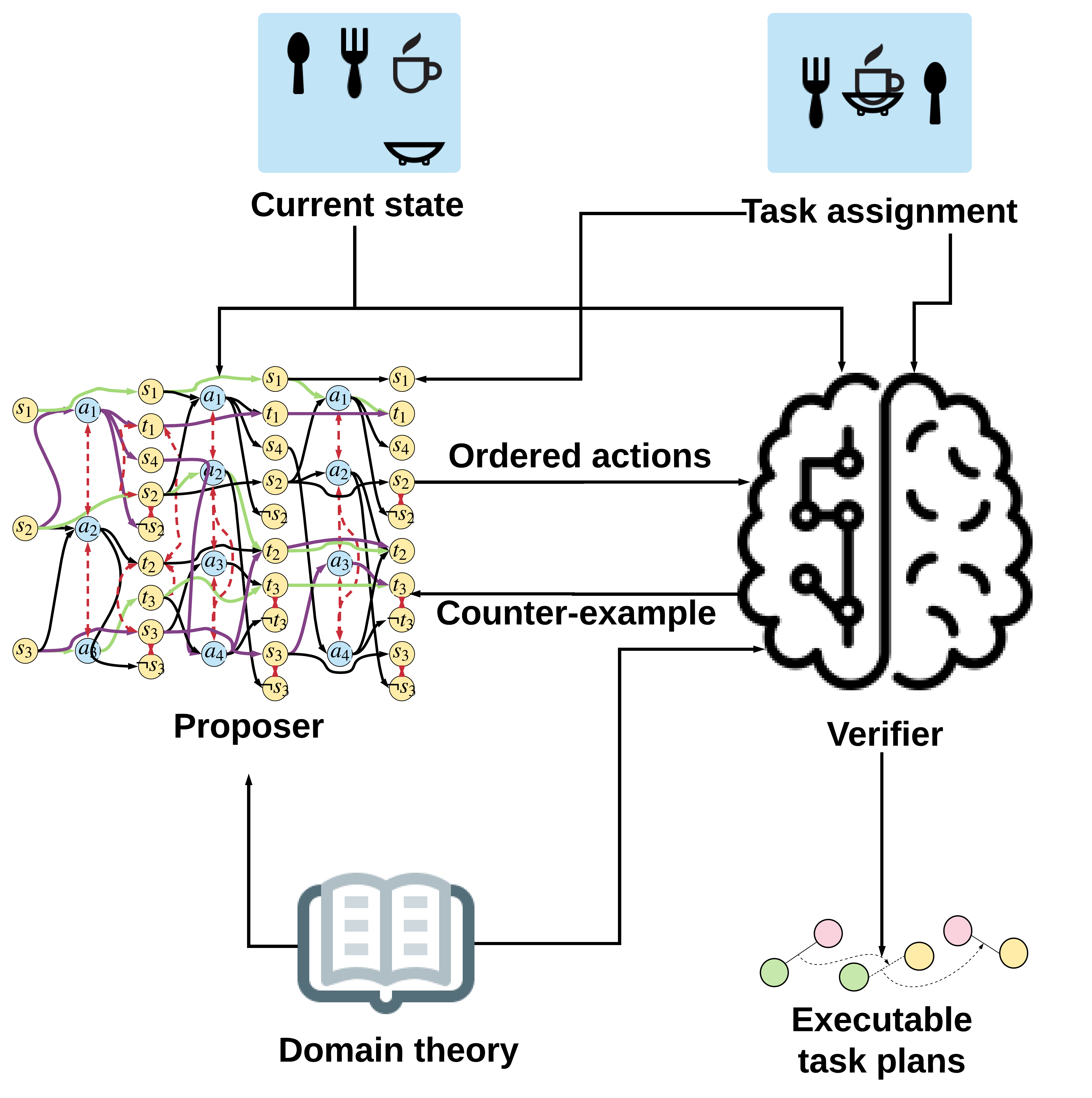}
    \caption{An overall framework for the automatic task planning with the proposer and the verifier}
    \label{framework}
\end{figure}

The overall framework is summarized in Fig. \ref{framework}. Given a parametric domain theory $a_1,a_2,\cdots$ obtained from the specification mining based on video, current spatial terms $\wedge_i s_i$, and a task assignment $\psi$, we aim to generate a detailed sequence of task plans such that the task assignment $\psi$ can be accomplished. 
In the framework, the proposer takes the current terms as the initial nodes and the task assignment $\psi$ as the target nodes. Available actions, along with preconditions and the effect of taking those actions, are obtained from the domain theory in (\ref{knowledge base}) and used in expanding the graph in the proposer. Ordered actions are generated by the backward solution extraction and passed to the verifier. The verifier takes the ordered actions from the proposer and verifies them based on the constraints posed by the domain theory, current spatial terms, and the task assignment. The verifier first checks the feasibility of temporal parameters in the parametric GSTL formulas from the proposer via SMT. Then it checks if there is a feasible solution for the spatial terms under logic constraints, which is solved by the SAT solver. If the actions are not executable, then it will inform the proposer that the current planning is infeasible. If the ordered actions are executable, they will be published for robots to implement. 

\section{Evaluation}\label{section:evaluation}

In this section, we evaluate the effectiveness of the proposed specification mining algorithm and automated task planning through a dining table setting example. We first generate a domain theory containing necessary information on solving the task, which is achieved by specification mining based on the video. Then, we perform automated task planning by implementing the proposer and the verifier introduced in the previous section.

\subsection{Specification mining}

\subsubsection{Data preprocessing}
We record several videos of table setting for the specification mining algorithm.
In order to obtain both color and depth images, we chose to use the ZED Stereo camera. The camera uses two lenses at a set distance apart to capture both a right and left color image for each frame. Using those images and the distance between the lenses, the camera's software is able to calculate depth measurements for each pixel of the frame. 

We first perform object detection on the obtained video. There are numerous results on object detection algorithms. As object detection is not the focus of this paper, we choose color-based filtering for object detection due to its robust performance.

\begin{figure}
    \centering
    \includegraphics[scale=0.1]{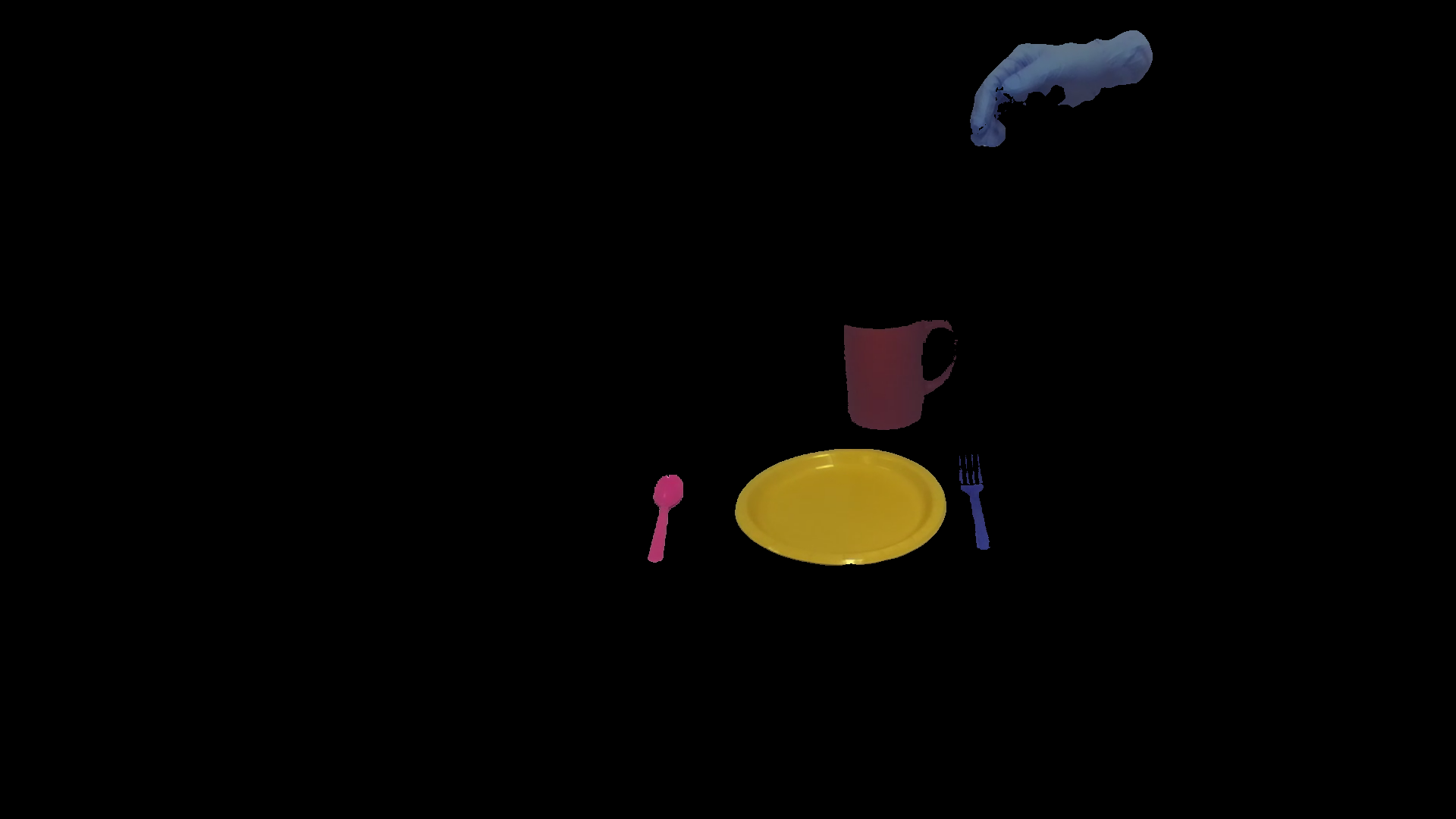}
    \caption{Color based object detection using HSV color scheme}
    \label{hsvfilter}
\end{figure}

The goal for object detection is to be able to isolate each object of interest individually and find a mask that can then be applied to the depth images and isolate each object's depth data. The first step to creating masks is color filtering. Each object in our test setup has a distinct color that will allow for isolation with a color filter. Hue, Saturation, Value (HSV) color scheme was used for the color filters, where hue is the base color or pigment, saturation is the amount of pigment, and value is the darkness. We further apply kernel filters and median filers to improve the performance of the color-based object detection by removing high-frequency noise. The usefulness of the HSV color scheme for color filtering is exemplified in Fig. \ref{hsvfilter}. 

\begin{figure}
    \centering
    \includegraphics[scale=0.1]{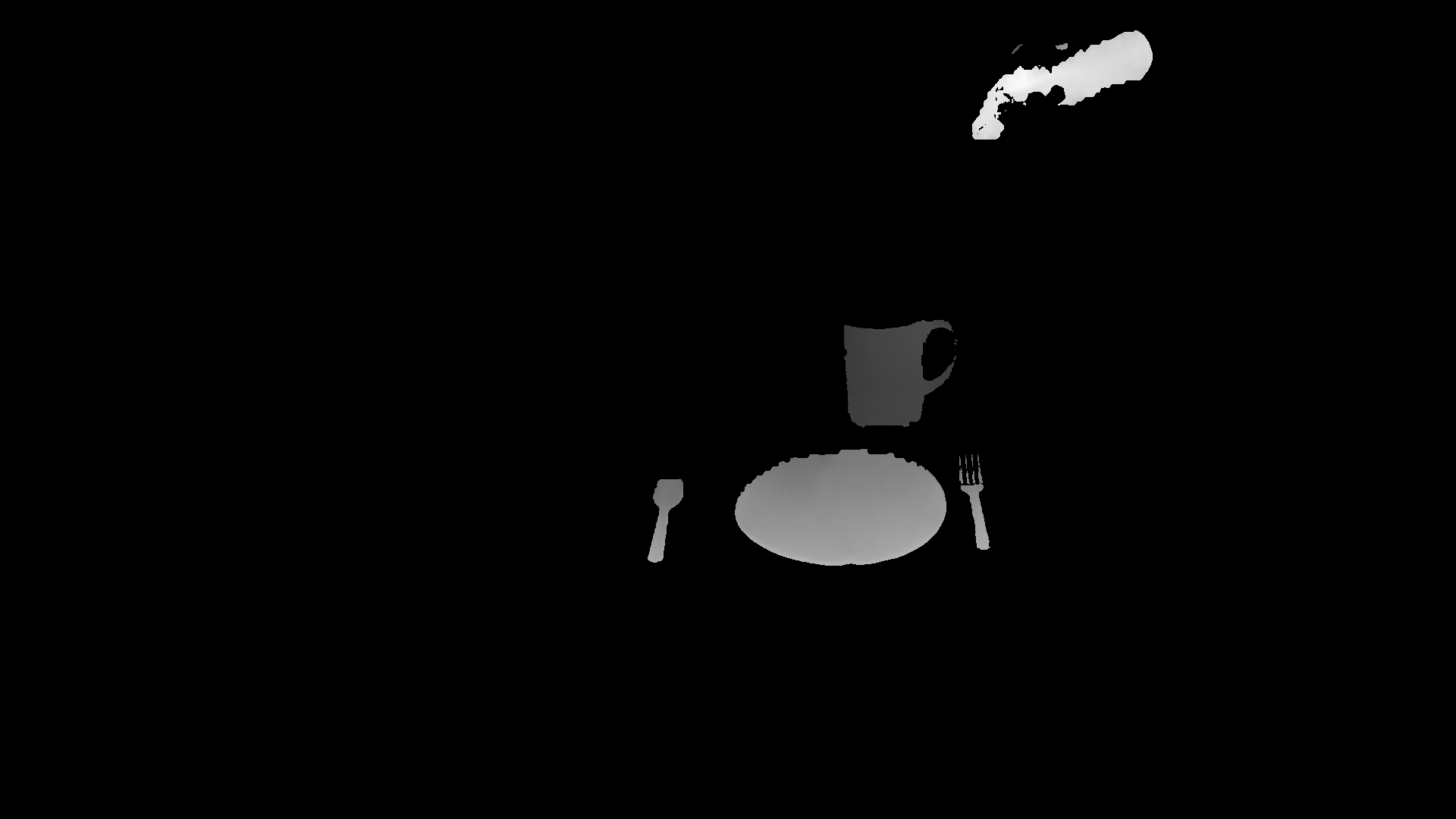}
    \caption{Depth}
    \label{depth information}
\end{figure}

Once the object masks are found, they can then be used to isolate each object in the corresponding depth image for each frame. With an object isolated in depth image, other parameters like average depth are calculated. This data is vital for specification mining because information like relative location and contact are very important to learn how the objects interact throughout a target process. Fig. \ref{depth information} shows the depth value for each object.

We consider six directional relations, namely front, back, left, right, top, and down, between any two objects that are closer than a certain distance.
For each frame, only objects with similar depth value are eligible for left, right, top, and down relations. Objects with similar horizontal positions in Fig. \ref{hsvfilter} and different depth values are eligible for front and back relations. We store the relative position between two objects in a table where each row records time, objects name, relative directional relations. The table will be used in the specification mining algorithm.

\subsubsection{Specification mining based on video}
We record six demo videos where a person set up the dining table using two different approaches. We first show detailed results for the first video and then overall results for all six videos. 
Following Algorithm \ref{specification mining algorithm}, we first generate spatial terms for each frame. We obtain the spatial terms $s_1$, $s_2$, $s_3$, $s^*_1$, $s^*_2$, $s^*_3$, and the  following spatial terms from the first video
\begin{align*}
    &a_1^{'}=\mathbf{C}_\exists^2(hand\wedge\mathbf{N}_\exists^{{\left \langle *, *, * \right \rangle}}cup)\\
    &a_3^{'}=\mathbf{C}_\exists^2(hand\wedge\mathbf{N}_\exists^{{\left \langle *, *, * \right \rangle}}spoon)\\
    &a_4^{'}=\mathbf{C}_\exists^2(hand\wedge\mathbf{N}_\exists^{{\left \langle *, *, * \right \rangle}}plate),\\
    &s_4=\mathbf{C}_\exists^2(fork\wedge\mathbf{N}_\exists^{left}empty).
\end{align*}
Some spatial terms are omitted for simplicity. After we merge the consecutive frame with the same spatial terms and mine ``Always" formulas, we have the following the ``Always" formulas for the first video
\begin{align*}
    &\Box_{[1,117]}s_1,~ \Box_{[1,75]}s_2,~ \Box_{[1,494]}s_3,~ \Box_{[118,339]}s_4\\
    &\Box_{[75,183]}a_1^{'},~ \Box_{[126,669]}s^*_1,~ \Box_{[274,386]}a_4^{'},\\ &\Box_{[340,669]}s^*_2,~ \Box_{[458,584]}a_3^{'},~ \Box_{[535,669]}s^*_3.
\end{align*}
Then, we learn temporal relations among these ``Always" formulas and obtain the following results for the video.
\begin{align*}
    &\Box_{[1,117]}s_1\sqcup_{[75,117]}^{o} \Box_{[75,183]}a_1^{'} \sqcup_{[126,183]}^o \Box_{[126,669]} s^*_1,\\
    &\Box_{[118,339]}s_4\sqcup_{[274,339]}^{o} \Box_{[274,386]}a_4^{'} \sqcup_{[340,386]}^o \Box_{[340,669]} s^*_2,\\
    &\Box_{[1,494]}s_3\sqcup_{[458,494]}^{o} \Box_{[458,584]}a_3^{'} \sqcup_{[535,584]}^o \Box_{[535,669]} s^*_3.
\end{align*}
After we apply the same algorithm to multiple video and replace the time instances with temporal variables, we obtain the following results.
\begin{align*}
    &a_1=(\Box_{[t_1,t_2]}s_1)\sqcup_{[\alpha_1,\beta_1]}^o (\Box_{[t_3,t_4]} a_1^{'}) \sqcup_{[\alpha_2,\beta_2]}^o(\Box_{[t_5,t_6]}s^*_1),~\\
    &a_2=(\Box_{[t_1,t_2]}s_2)\sqcup_{[\alpha_1,\beta_1]}^o (\Box_{[t_3,t_4]} a_2^{'}) \sqcup_{[\alpha_2,\beta_2]}^o(\Box_{[t_5,t_6]}s^*_2),~\\
    &a_3=(\Box_{[t_1,t_2]}s_3)\sqcup_{[\alpha_1,\beta_1]}^o (\Box_{[t_3,t_4]} a_3^{'})\sqcup_{[\alpha_2,\beta_2]}^o(\Box_{[t_5,t_6]}s^*_3),~\\
    &a_4=(\Box_{[t_1,t_2]}s_4)\sqcup_{[\alpha_1,\beta_1]}^o (\Box_{[t_3,t_4]} a_4^{'})\sqcup_{[\alpha_2,\beta_2]}^o(\Box_{[t_5,t_6]}s^*_2),
\end{align*}
where $a_2^{'}=\mathbf{C}_\exists^2(hand\wedge\mathbf{N}_\exists^{{\left \langle *, *, * \right \rangle}}fork)$ is the spatial term learned from other video.
\subsection{Automated task planning}

\subsubsection{Proposer}
\begin{figure*}
 \centering
 \includegraphics[scale=0.25]{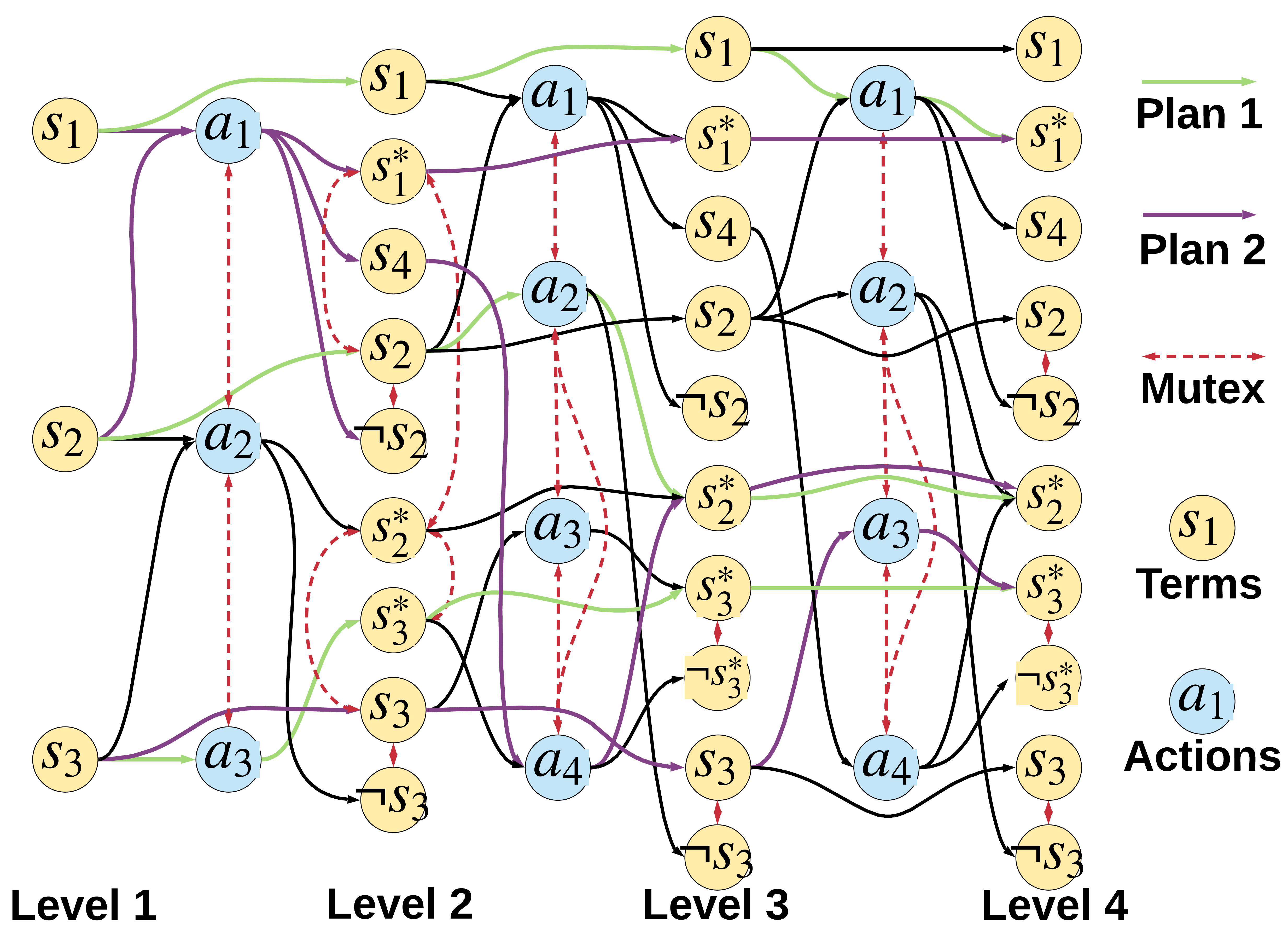}
 \caption{Automatic task planning based on forward graph expansion and backward solution extraction.}
 \label{exp:task planning}
 %\vspace{-4mm}
\end{figure*}

We use the same example in Fig. \ref{exp:task planning example} to evaluate the proposer. Fig. \ref{exp:task planning} illustrates the graph expanding and solution extraction algorithm. Initially, we have three terms $s_1,s_2,\text{and}~s_3$ and three available actions $a_1,a_2,\text{and}~a_3$. We expand the graph by generating terms $s^*_1,s_4,\text{and}~\neg s_2$ from applying $a_1$ to $s_1$, terms $s^*_2$ and $\neg s_3$ from applying $a_2$ to $s_2$, and terms $s^*_3$ from applying $a_3$ to $s_3$. $a_1$ and $a_2$ are mutex since $a_1$ generate the negation of the precondition of $a_2$. $a_2$ and $a_3$ are mutex for the same reason. Consequently, $s^*_1$ and $s^*_2$ are mutex since their supporting actions $a_1$ and $a_2$ are mutex. We label all mutex relations as the red dash line in Fig. \ref{exp:task planning}. Even though the current term level includes all target terms, we need to further expand the graph as $s^*_1$ and $s^*_2$ are mutex. Notice that we move terms to the next level if there are no actions applied. We expand the second term level following the same procedure and label all mutex with red dash lines. In the third term level, $s^*_1$ and $s^*_2$ are not mutex anymore since they both have ``no action" as their supporting actions, and they are not mutex. Since the third term level includes all target terms, and there is no mutex, we now move to the backward solution extraction.
 
From the previous forward graph expansion phase, we obtain a graph with three levels of terms and two levels of actions. From Fig. \ref{exp:task planning}, we can see that available actions for $s^*_1$ at level 3 is $\mathcal{A}_{{s^*_1}^3}=\{a_1,\varnothing\}$, available actions for $s^*_2$ is $\mathcal{A}_{{s^*_2}^3}=\{a_2,\varnothing,a_4\}$, and available actions for $s^*_3$ is $\mathcal{A}_{{s^*_3}^3}=\{a_3,\varnothing\}$, where $a_1$ and $a_2$ are mutex, $a_2$ and $a_3$ are mutex, $a_2$ and $a_4$, and $a_3$ and $a_4$ are mutex. Thus, all possible solutions for action level 2 are $\{\{a_1,\varnothing,a_3\}$, $\{a_1,\varnothing,\varnothing\}$, $\{a_1,a_4,\varnothing\}$, $\{\varnothing,a_2,\varnothing\}$, $\{\varnothing,\varnothing,a_3\}$, $\{\varnothing,\varnothing,\varnothing\}$, $\{\varnothing,a_4,\varnothing\}\}$. Let us take $\mathcal{A}^k=\{a_1,\varnothing,a_3\}$ as an example. The precondition for it at level 2 is $\{s_1,s^*_2,s_3\}$. Since $s_3$ and $t_2$ are mutex, thus $\mathcal{A}^k=\{a_1,\varnothing,a_3\}$ is not a feasible solution. In fact, there are no feasible solution for the current transition system. In the case where no feasible solution after the solution extraction phase enumerate all possible candidate, we go back to the graph expanding phase and further grow the graph. For example in Fig. \ref{exp:task planning}, after we grow another level of actions and terms, the solution extraction phase is able to find two possible ordered actions: $a_3,a_2,a_1$ and $a_1,a_4,a_3$. They are highlighted with green and purple lines in Fig. \ref{exp:task planning} respectively. The results will be tested in the verifier to make sure the plan is executable.

\subsubsection{Verifier}
Let us continue the dining table set up example. We write one of the ordered actions given by the proposer as a GSTL formula $\phi=a'_3\sqcup_{[e_1,e_2]}^ba'_2\sqcup_{[e_3,e_4]}^ba'_1$. We assume the domain theory requires that robots need 5 seconds to move spoon, fork, and cup. The task assignment is setting up the table in 40 seconds which can be represented as a GSTL formula $\psi=\Diamond_{[0,40]}(s^*_1\wedge s^*_2\wedge s^*_3)$. Based on the domain theory we obtained from the specification mining algorithm, we have the following parametric GSTL formulas.
\begin{align*}
    &a_i=(\Box_{[t_{i,0},t_{i,0}+c_i+5]}s_i)\sqcup_{[t_{i,0}+c_i,t_{i,0}+c_i+5]}^o(\Box_{[t_{i,0}+c_i,t_{i,0}+d_i+5]} a_i^{'})\\
    &\sqcup_{[t_{i,0}+d_i,t_{i,0}+d_i+5]}^o(\Box_{[t_{i,0}+d_i,t_{i,0}+d_i+5+\epsilon]}s^*_i),~\forall i\in [1,2,3],\\
    %&a_3=(\Box_{[t_{3,0},t_{3,0}+c_3+5]}s_3)\sqcup_{[t_0+a,t_0+c_3+5]}^o(\Box_{[t_0+a,t_0+d_3+5]} a_3^{'})\\
    %&\sqcup_{[t_0+b,t_0+b+5]}^o(\Box_{[t_0+b,t_0+b+5+\epsilon]}t_3),\\
    %&a_2=(\Box_{[t_0,t_0+a+5]}s_2)\sqcup_{[t_0+a,t_0+a+5]}^o(\Box_{[t_0+a,t_0+b+5]} a_2^{'})\\
    %&\sqcup_{[t_0+b,t_0+b+5]}^o(\Box_{[t_0+b,t_0+b+5+\epsilon]}t_2),\\
    %&a_1=(\Box_{[t_0,t_0+a+5]}s_1)\sqcup_{[t_0+a,t_0+a+5]}^o(\Box_{[t_0+a,t_0+b+5]} a_1^{'})\\
    %&\sqcup_{[t_0+b,t_0+b+5]}^o(\Box_{[t_0+b,t_0+b+5+\epsilon]}t_1),\\
    &\phi=a'_3\sqcup_{[e_1,e_2]}^ba'_2\sqcup_{[e_3,e_4]}^ba'_1,\\
    &\psi=\Diamond_{[0,40]}(s^*_1\wedge s^*_2\wedge s^*_3).
\end{align*}
The job for the verifier is to find a set of value for $c_i$, $d_i$, $e_i$ and $t_{i,0}$ for each action such that $\psi$ is satisfied and no temporal constraint is violated. Using the proposed algorithm, we first reformulate the GSTL formula in $\wedge_i(\vee_j f_{i,j})$ form, where $f_{i,j}$ is the linear inequality predicate for the temporal parameters. We obtain the following SMT encoding for the parametric GSTL formulas above
\begin{equation}
    \begin{aligned}
    &\forall i \in [1,2,3],\\
    &t_{s_i}^1\leq t_{i,0},~ t_{i,0}+c_i+5\leq t_{s_i}^2,\\
    &t_{a_{i}^{'}}^1\leq t_{i,0}+c_i,~ t_{i,0}+c_i+5\leq t_{a_{i}^{'}}^2,\\
    &t_{a_{i}^{'}}^1\leq t_{i,0}+d_i,~ t_{i,0}+d_i+5\leq t_{a_{i}^{'}}^2,\\
    &t_{s^*_i}^1\leq t_{i,0}+d_i,~ t_{i,0}+d_i+5+\epsilon\leq t_{s^*_i}^2,\\
    &t_{a'_3}^2\leq e_1,~ e_2\leq t_{a'_2}^1,~ t_{a'_2}^2\leq e_3,~ e_4\leq t_{a'_1}^1,\\
    &t_{s^*_i}^1\leq 40,
    \end{aligned}
\end{equation}
where $t_{\tau}^1$ and $t_{\tau}^2$ is the lower bound and upper bound of spatial term $\tau$. All constraints are connected with conjunction operators. We employ the SMT solver MathSAT \cite{mathsat5} and implemented in Python with pySMT API \cite{pysmt2015}. The SMT solver returns the following results.
\begin{equation}
    \begin{aligned}
    &\phi=a'_3\sqcup_{[13,14]}^ba'_2\sqcup_{[25,26]}^ba'_1\\
    &a_3=(\Box_{[1,7]}s_3)\sqcup_{[2,7]}^o(\Box_{[2,13]}a'_3)\sqcup_{[8,13]}^o\Box_{[8,14]}s^*_3,\\
    &a_2=(\Box_{[13,19]}s_2)\sqcup_{[14,19]}^o(\Box_{[14,25]}a'_2)\sqcup_{[20,25]}^o\Box_{[20,26]}s^*_2,\\
    &a_1=\Box_{[25,31]}s_1\sqcup_{[26,31]}^o(\Box_{[26,37]}a'_1)\sqcup_{[32,37]}^o\Box_{[32,38]}s^*_1.
\end{aligned}
\label{smt result}
\end{equation}

As we can see from the above CNF form \eqref{smt result}, temporal parameters are solved by the SMT solver. To verify the spatial terms in \eqref{smt result}, we use the Boolean encoding in \cite{liu2020graph} and obtain the following CNF form for $s_3$ as an example.
\begin{equation}
\begin{aligned}
    &s_3=\mathbf{C}_\exists^2(spoon\wedge\mathbf{N}_\exists^{left}fork)
    =\bigvee_{j=1}^{n_j}\left(\varphi_j\wedge\phi_j\right)\\
    &=\bigwedge\begin{pmatrix}
\varphi_1\vee\phi_1 & \varphi_1\vee\phi_2 &... &\varphi_1\vee\phi_{n_j}\\ 
\varphi_2\vee\phi_1 & \varphi_2\vee\phi_2 &... &\varphi_2\vee\phi_{n_j} \\ 
 \vdots&\vdots  & ...&\vdots\\ 
\varphi_{n_j}\vee\phi_1 & \varphi_{n_j}\vee\phi_2 &... &\varphi_{n_j}\vee\phi_{n_j} 
\end{pmatrix}\\
& \varphi_j=\bigvee_{i=1}^n\mathbf{C}_{A_j}\mathbf{C}_{A_i}spoon,\\
&\phi_j=\bigvee_{i=1}^{n_i}\bigvee_{k=1}^{n_k}\mathbf{C}_{A_j}\mathbf{C}_{A_i}\mathbf{N}_{A_k}^{left}fork,
\end{aligned}
\label{CNF2}
\end{equation}
where the truth value of $\varphi_j$ and $\phi_j$ are to be assigned by the SAT solver. We apply the same procedure for the rest of spatial term in \eqref{smt result} at different time and use a SAT solver to check the feasibility of the set of obtained logic constraints. We use PicoSAT \cite{biere2008picosat} as the SAT solver where each spatial term at a different time is modeled as a Boolean variable. The solver returns a feasible solution meaning the task plans generated by the proposer is feasible. 

Let us assume the domain theory requires that robots need 10 seconds to move a plate with the following GSTL formulas
\begin{align*}
    &a_4=\Box_{[t_{4,0},t_{4,0}+c_4+10]}s_4\sqcup_{[t_{4,0}+c_4,t_{4,0}+c_4+10]}^o(\Box_{[t_{4,0}+c_4,t_{4,0}+d_4+10]}a'_4)\\
    &\sqcup_{[t_{4,0}+d_4,t_{4,0}+d_4+10]}^o\Box_{[t_{4,0}+d_4,t_{4,0}+d_4+10+\epsilon]}s^*_2.
\end{align*}
We use the same algorithm for the verifier to check the feasibility of the ordered actions from the proposer. The verifier cannot find a feasible solution for the other ordered actions $\phi=a'_1\sqcup_{[e_1,e_2]}^ba'_4\sqcup_{[e_3,e_4]}^ba'_3$ because the SMT cannot find a feasible solution where the task assignment can be accomplished within 40 seconds.

\section{Conclusion}\label{Section: conclusion}
We study specification mining based on demo videos and automated task planning for autonomous robots using GSTL. We use GSTL formulas to represent spatial and temporal information for autonomous robots. We generate the domain theory in GSTL by learning from demo videos and use the domain theory in the automatic task planning.
An automatic task planning framework is proposed with an interacted proposer and verifier. The proposer generates ordered actions with unknown temporal parameters by running the graph expansion phase and the solution extraction phase iteratively. The verifier verifies if the plan is feasible and outputs executable task plans through an SMT solver for temporal feasibility and an SAT solver for spatial feasibility.

\begin{comment}

\begin{equation}
\begin{aligned}
    &\left(\bigwedge^{t_0+a}_{i=t_0}s_3\right) 
    \left(\bigwedge^{t_0+a+5}_{i=t_0+a}(s_3\wedge (hand\wedge\mathbf{N}_\exists^{\left \langle *, *, * \right \rangle}spoon)\right)\\
    &\left(\bigwedge^{t_0+b+5}_{i=t_0+b}t_3\wedge (hand\wedge\mathbf{N}_\exists^{\left \langle *, *, * \right \rangle}spoon)\right)
    \left(\bigwedge^{t_0+b+5+\epsilon}_{i=t_0+b+5}t_3\right).
\end{aligned}
\label{CNF1}
\end{equation}

One of the feasible solution given by the solver could be 
\begin{align*}
    &\phi=a_3\sqcup_{[14,15]}^ba_2\sqcup_{[30,31]}^ba_1\\
    &a_3=\Box_{[0,6]}s_3\sqcup_{[1,6]}^o\mathbf{C}_\exists^2(hand\wedge\mathbf{N}_\exists^{\left \langle *, *, * \right \rangle}spoon)\sqcup_{[7,12]}^o\Box_{[7,13]}t_3,\\
    &a_2=\Box_{[16,22]}s_2\sqcup_{[17,22]}^o\mathbf{C}_\exists^2(hand\wedge\mathbf{N}_\exists^{\left \langle *, *, * \right \rangle}fork)\sqcup_{[23,28]}^o\Box_{[23,29]}t_2,\\
    &a_1=\Box_{[32,38]}s_1\sqcup_{[33,38]}^o\mathbf{C}_\exists^2(hand\wedge\mathbf{N}_\exists^{\left \langle *, *, * \right \rangle}cup)\sqcup_{[39,42]}^o\Box_{[39,45]}t_1.
\end{align*}

\end{comment}

%\section*{References}

\bibliography{mybibfile}

\begin{thebibliography}{10}
\expandafter\ifx\csname url\endcsname\relax
  \def\url#1{\texttt{#1}}\fi
\expandafter\ifx\csname urlprefix\endcsname\relax\def\urlprefix{URL }\fi
\expandafter\ifx\csname href\endcsname\relax
  \def\href#1#2{#2} \def\path#1{#1}\fi

\bibitem{paulius2019survey}
D.~Paulius, Y.~Sun, A survey of knowledge representation in service robotics,
  Robotics and Autonomous Systems 118 (2019) 13--30.

\bibitem{liu2020graph}
Z.~Liu, M.~Jiang, H.~Lin, A graph-based spatial temporal logic for knowledge
  representation and automated reasoning in cognitive robots, arXiv preprint
  arXiv:2001.07205.

\bibitem{kong2017temporal}
Z.~Kong, A.~Jones, C.~Belta, Temporal logics for learning and detection of
  anomalous behavior, IEEE Transactions on Automatic Control 62~(3) (2017)
  1210--1222.

\bibitem{nenzi2018robust}
L.~Nenzi, S.~Silvetti, E.~Bartocci, L.~Bortolussi, A robust genetic algorithm
  for learning temporal specifications from data, in: International Conference
  on Quantitative Evaluation of Systems, Springer, 2018, pp. 323--338.

\bibitem{bombara2016decision}
G.~Bombara, C.-I. Vasile, F.~Penedo, H.~Yasuoka, C.~Belta, A decision tree
  approach to data classification using signal temporal logic, in: Proceedings
  of the 19th International Conference on Hybrid Systems: Computation and
  Control, ACM, 2016, pp. 1--10.

\bibitem{jin2015mining}
X.~Jin, A.~Donz{\'e}, J.~V. Deshmukh, S.~A. Seshia, Mining requirements from
  closed-loop control models, IEEE Transactions on Computer-Aided Design of
  Integrated Circuits and Systems 34~(11) (2015) 1704--1717.

\bibitem{bartocci2016formal}
E.~Bartocci, E.~A. Gol, I.~Haghighi, C.~Belta, A formal methods approach to
  pattern recognition and synthesis in reaction diffusion networks, IEEE
  Transactions on Control of Network Systems 5~(1) (2016) 308--320.

\bibitem{weld1999recent}
D.~S. Weld, Recent advances in ai planning, AI magazine 20~(2) (1999) 93--93.

\bibitem{li2012planning}
Y.~Li, J.~Sun, J.~S. Dong, Y.~Liu, J.~Sun, Planning as model checking tasks,
  in: 2012 35th Annual IEEE Software Engineering Workshop, IEEE, 2012, pp.
  177--186.

\bibitem{zhou2018mobile}
Z.~Zhou, J.~Feng, B.~Gu, B.~Ai, S.~Mumtaz, J.~Rodriguez, M.~Guizani, When
  mobile crowd sensing meets uav: Energy-efficient task assignment and route
  planning, IEEE Transactions on Communications 66~(11) (2018) 5526--5538.

\bibitem{zheng2019vector}
W.~Zheng, H.~Lin, Vector autoregressive pomdp model learning and planning for
  human--robot collaboration, IEEE Control Systems Letters 3~(3) (2019)
  775--780.

\bibitem{wachter2018integrating}
M.~W{\"a}chter, E.~Ovchinnikova, V.~Wittenbeck, P.~Kaiser, S.~Szedmak,
  W.~Mustafa, D.~Kraft, N.~Kr{\"u}ger, J.~Piater, T.~Asfour, Integrating
  multi-purpose natural language understanding, robot’s memory, and symbolic
  planning for task execution in humanoid robots, Robotics and Autonomous
  Systems 99 (2018) 148--165.

\bibitem{hertzberg2008ai}
J.~Hertzberg, R.~Chatila, Ai reasoning methods for robotics, Springer handbook
  of robotics (2008) 207--223.

\bibitem{post1921introduction}
E.~L. Post, Introduction to a general theory of elementary propositions,
  American journal of mathematics 43~(3) (1921) 163--185.

\bibitem{mccarthy1960programs}
J.~McCarthy, Programs with common sense, RLE and MIT computation center, 1960.

\bibitem{baader2003description}
F.~Baader, D.~Calvanese, D.~McGuinness, P.~Patel-Schneider, D.~Nardi, The
  description logic handbook: Theory, implementation and applications,
  Cambridge university press, 2003.

\bibitem{cohn2001qualitative}
A.~G. Cohn, S.~M. Hazarika, Qualitative spatial representation and reasoning:
  An overview, Fundamenta informaticae 46~(1-2) (2001) 1--29.

\bibitem{raman2015reactive}
V.~Raman, A.~Donz{\'e}, D.~Sadigh, R.~M. Murray, S.~A. Seshia, Reactive
  synthesis from signal temporal logic specifications, in: Proceedings of the
  18th International Conference on Hybrid Systems: Computation and Control
  (HSCC), ACM, 2015, pp. 239--248.

\bibitem{kontchakov2007spatial}
R.~Kontchakov, A.~Kurucz, F.~Wolter, M.~Zakharyaschev, Spatial logic+ temporal
  logic=?, in: Handbook of spatial logics, Springer, 2007, pp. 497--564.

\bibitem{haghighi2016robotic}
I.~Haghighi, S.~Sadraddini, C.~Belta, Robotic swarm control from
  spatio-temporal specifications, in: 2016 IEEE 55th Conference on Decision and
  Control (CDC), IEEE, 2016, pp. 5708--5713.

\bibitem{bartocci2017monitoring}
E.~Bartocci, L.~Bortolussi, M.~Loreti, L.~Nenzi, Monitoring mobile and
  spatially distributed cyber-physical systems, in: Proceedings of the 15th
  ACM-IEEE International Conference on Formal Methods and Models for System
  Design, ACM, 2017, pp. 146--155.

\bibitem{suomalainen2017geometric}
M.~Suomalainen, V.~Kyrki, A geometric approach for learning compliant motions
  from demonstration, in: 2017 IEEE-RAS 17th International Conference on
  Humanoid Robotics (Humanoids), IEEE, 2017, pp. 783--790.

\bibitem{decker2017service}
M.~Decker, M.~Fischer, I.~Ott, Service robotics and human labor: A first
  technology assessment of substitution and cooperation, Robotics and
  Autonomous Systems 87 (2017) 348--354.

\bibitem{argall2009survey}
B.~D. Argall, S.~Chernova, M.~Veloso, B.~Browning, A survey of robot learning
  from demonstration, Robotics and autonomous systems 57~(5) (2009) 469--483.

\bibitem{tsurumine2019deep}
Y.~Tsurumine, Y.~Cui, E.~Uchibe, T.~Matsubara, Deep reinforcement learning with
  smooth policy update: Application to robotic cloth manipulation, Robotics and
  Autonomous Systems 112 (2019) 72--83.

\bibitem{sutton2018reinforcement}
R.~S. Sutton, A.~G. Barto, Reinforcement learning: An introduction, MIT press,
  2018.

\bibitem{asarin2011parametric}
E.~Asarin, A.~Donz{\'e}, O.~Maler, D.~Nickovic, Parametric identification of
  temporal properties, in: International Conference on Runtime Verification,
  Springer, 2011, pp. 147--160.

\bibitem{bartocci2013robustness}
E.~Bartocci, L.~Bortolussi, L.~Nenzi, G.~Sanguinetti, On the robustness of
  temporal properties for stochastic models, arXiv preprint arXiv:1309.0866.

\bibitem{yang2012querying}
H.~Yang, B.~Hoxha, G.~Fainekos, Querying parametric temporal logic properties
  on embedded systems, in: IFIP International Conference on Testing Software
  and Systems, Springer, 2012, pp. 136--151.

\bibitem{hawes2017strands}
N.~Hawes, C.~Burbridge, F.~Jovan, L.~Kunze, B.~Lacerda, L.~Mudrova, J.~Young,
  J.~Wyatt, D.~Hebesberger, T.~Kortner, et~al., The strands project: Long-term
  autonomy in everyday environments, IEEE Robotics \& Automation Magazine
  24~(3) (2017) 146--156.

\bibitem{veloso2012cobots}
M.~Veloso, J.~Biswas, B.~Coltin, S.~Rosenthal, T.~Kollar, C.~Mericli,
  M.~Samadi, S.~Brandao, R.~Ventura, Cobots: Collaborative robots servicing
  multi-floor buildings, in: 2012 IEEE/RSJ international conference on
  intelligent robots and systems, IEEE, 2012, pp. 5446--5447.

\bibitem{tran2017robots}
T.~T. Tran, T.~Vaquero, G.~Nejat, J.~C. Beck, Robots in retirement homes:
  Applying off-the-shelf planning and scheduling to a team of assistive robots,
  Journal of Artificial Intelligence Research 58 (2017) 523--590.

\bibitem{kunze2018artificial}
L.~Kunze, N.~Hawes, T.~Duckett, M.~Hanheide, T.~Krajn{\'\i}k, Artificial
  intelligence for long-term robot autonomy: A survey, IEEE Robotics and
  Automation Letters 3~(4) (2018) 4023--4030.

\bibitem{zhao2019object}
Z.-Q. Zhao, P.~Zheng, S.-t. Xu, X.~Wu, Object detection with deep learning: A
  review, IEEE transactions on neural networks and learning systems 30~(11)
  (2019) 3212--3232.

\bibitem{chaudhary2018learning}
K.~Chaudhary, K.~Wada, X.~Chen, K.~Kimura, K.~Okada, M.~Inaba, Learning to
  segment generic handheld objects using class-agnostic deep comparison and
  segmentation network, IEEE Robotics and Automation Letters 3~(4) (2018)
  3844--3851.

\bibitem{allen1983maintaining}
J.~F. Allen, Maintaining knowledge about temporal intervals, Communications of
  the ACM 26~(11) (1983) 832--843.

\bibitem{lifschitz2008knowledge}
V.~Lifschitz, L.~Morgenstern, D.~Plaisted, Knowledge representation and
  classical logic, Foundations of Artificial Intelligence 3 (2008) 3--88.

\bibitem{amir2005partition}
E.~Amir, S.~McIlraith, Partition-based logical reasoning for first-order and
  propositional theories, Artificial intelligence 162~(1-2) (2005) 49--88.

\bibitem{mathsat5}
A.~Cimatti, A.~Griggio, B.~Schaafsma, R.~Sebastiani, {The MathSAT5 SMT Solver},
  in: N.~Piterman, S.~Smolka (Eds.), Proceedings of TACAS, Vol. 7795 of LNCS,
  Springer, 2013.

\bibitem{pysmt2015}
M.~Gario, A.~Micheli, Pysmt: a solver-agnostic library for fast prototyping of
  smt-based algorithms, in: SMT Workshop 2015, 2015.

\bibitem{biere2008picosat}
A.~Biere, Picosat essentials, Journal on Satisfiability, Boolean Modeling and
  Computation 4~(2-4) (2008) 75--97.

\end{thebibliography}

\end{document}